%% file: iclr2026_conference.tex
\documentclass{article} 

\usepackage{iclr2026_conference,times}

\usepackage{times}

\input{math_commands.tex}

\usepackage{url}

\usepackage{graphicx}
\usepackage{subcaption} 

\usepackage{amsmath}
\usepackage{amssymb}
\usepackage{amsthm}

\usepackage{booktabs} 
\usepackage{tabularx} 
\usepackage{array}

\usepackage{xcolor}
\usepackage{soul}

\theoremstyle{plain} 

\theoremstyle{definition} 

\theoremstyle{remark} 

\usepackage{enumitem}
\usepackage{booktabs}
\usepackage{multirow}

\usepackage{booktabs, multirow, makecell, xcolor, caption, adjustbox}

\usepackage{makecell}
\usepackage{siunitx}
 \usepackage{array}
 \newcolumntype{Z}{>{\hsize=1.25\hsize\raggedright\arraybackslash}X}  
 \newcolumntype{Y}{>{\hsize=0.75\hsize\raggedright\arraybackslash}X} 

\makeatletter
\def\iclrheader#1{} 

\renewcommand{\@maketitle}{
  \vbox{
    \begin{flushleft}
    \end{flushleft}
    \vskip -0.4in
    \hrule height 1.5pt
    \vskip 0.25in
    \begin{center}
      {\LARGE\bf \@title \par}
    \end{center}
    \vskip 0.15in
    \hrule height 0.8pt
    \vskip 0.2in
    \begin{center}
      {\large \@author}
    \end{center}
    \vskip 0.25in
  }
}
\makeatother

\title{Don't Pay Attention}

\iclrfinalcopy 

\usepackage[utf8]{inputenc}
\usepackage[T1]{fontenc}
\usepackage{lmodern}
\usepackage{geometry}
\usepackage{xcolor}
\usepackage[most]{tcolorbox}
\usepackage{svg}            
\usepackage{hyperref}
\hypersetup{colorlinks=true, urlcolor=blue}
\usepackage{microtype}

\svgpath{{svg-inkscape/}{.}}

\hypersetup{
  colorlinks=false,                 
  citebordercolor={0.1 1 0.1},
  linkbordercolor={0.902 0.984 1.000},          
  urlbordercolor={0.902 0.984 1.000},           
  pdfborderstyle={/S/S/W 1}         
}

\let\Oldhref\href
\renewcommand{\href}[2]{\Oldhref{#1}{\textcolor{blue}{#2}}}
\let\Oldurl\url
\renewcommand{\url}[1]{\textcolor{blue}{\Oldurl{#1}}}

\definecolor{AveyPanel}{HTML}{E6FBFF} 
\definecolor{AveyText}{HTML}{333333}  

\newtcolorbox{aveybox}[1][]{%
  enhanced,
  colback=AveyPanel,
  colframe=AveyPanel,
  arc=12pt,
  boxrule=0pt,
  left=12pt,right=12pt,top=24pt,bottom=12pt,
  overlay={%
    \node[anchor=north east, inner sep=12pt]
      at (frame.north east)
      {\includesvg[width=1.8cm]{Avey_logo}}; 
  },
  #1
}

\newcommand{\authorentry}[3]{%
  \textbf{#1},\; #2,\; \href{mailto:#3}{\textcolor{AveyText}{#3}}%
}

\newcommand{\aveyheader}[4]{%
  \begingroup
  \color{AveyText}%
  \begin{aveybox}
    {\sffamily\bfseries\Large #1}\par\vspace{8pt}
    {\sffamily
      \authorentry{Mohammad Hammoud$^{*}$}{Avey AI}{mhh@avey.ai}\par
      \authorentry{Devang Acharya$^{*}$}{Avey AI}{dacharya@avey.ai}\par
    }\vspace{10pt}
    {\sffamily\small #4}\par\vspace{10pt}
    {\sffamily\scriptsize #3}%
    
    \vspace{6pt}
    \begin{flushleft}
      {\sffamily\scriptsize $^{*}$Equal contribution.}
    \end{flushleft}
  \end{aveybox}
  \endgroup
}

\usepackage{fancyhdr} 
\AtBeginDocument{%
  \fancyhf{}
  \renewcommand{\headrulewidth}{0pt}
  \fancyfoot[C]{\thepage}
}

\begin{document}

\aveyheader
  {Don't Pay Attention}
  {} 
  {Date: \today \quad Website: \href{https://avey.ai/research}{https://avey.ai/research}}
  {The Transformer has become the de facto standard for modern language models owing to its parallelizable training and effective autoregressive decoding. However, its fixed context window and the quadratic time and memory costs of its self-attention mechanism remain central bottlenecks. These constraints have revived interest in recurrent architectures that scale linearly with sequence length, but at the cost of reduced parallelism. In this paper, we introduce Avey, a new foundational architecture that breaks away from both attention and recurrence. Avey pairs a ranker with an autoregressive neural processor to select and contextualize only the most relevant tokens for any given token. Specifically, it decouples sequence length from context width, thus enabling effective and efficient processing of arbitrarily long sequences. Results show that Avey compares favorably to the Transformer across a variety of standard short-range NLP benchmarks, while significantly outperforming it on tasks requiring long-range dependency modeling.}

\section{Introduction}
\label{intro}

The Transformer~\citep{vaswani2017attention} has emerged as one of the most influential AI innovations in recent years, profoundly impacting various aspects of modern life, including work, science, and art, to mention just a few. Notably, Large Language Models (LLMs) are almost universally based on the Transformer~\citep{gu2023mamba}, which has demonstrated remarkable performance in natural language processing (NLP)~\citep{ouyang2022training, liu2019roberta, raffel2020exploring} and various other fields~\citep{he2022masked, liu2021swin, baevski2020wav2vec}.

The Transformer’s state-of-the-art performance is largely driven by its recurrence-free self-attention mechanism, which enables parallel processing of entire token sequences. Nevertheless, the computational and memory costs of self-attention grow quadratically with sequence length, making it inefficient for handling arbitrarily long sequences. Extensive research has been conducted over the years to address this limitation~\citep{tay2022efficient}, with a noticeable emphasis on linearizing attention~\citep{katharopoulos2020transformers, choromanski2020rethinking, zhai2021attention, wang2020linformer}. These linear approaches aim at approximating self-attention in a more computationally efficient manner, without considerably compromising performance.

Nonetheless, linear attention mechanisms have generally underperformed the original self-attention mechanism, often by a significant margin in language modeling tasks~\citep{yang2023gated, kasai2021finetuning}. While recent linear models such as RWKV~\citep{peng2025rwkv} and RetNet~\citep{sun2023retentive} have shown promising results, substantial progress is still needed before they can reliably and consistently surpass the Transformer~\citep{li2024survey}. In addition, these models have yet to demonstrate definitive empirical effectiveness at scale~\citep{gu2023mamba}. This persistent performance gap between quadratic and linear approaches has spurred renewed interest in RNN-based architectures, which offer linear scalability with sequence length but limit parallelism due to their inherently cyclical nature.

To exemplify, state space models (SSMs)~\citep{kalman1960new, gu2021combining}, which are viewed as extensions of RNNs, have recently emerged as a compelling class of architectures. Unlike traditional RNNs, SSMs can parameterize their state transition matrices in a structured manner (e.g., via using a diagonal plus low-rank decomposition) to improve computational efficiency and enhance gradient flow. A specialized subclass of these models, known as structured state space sequence (S4) models~\citep{gu2021efficiently, gu2021combining}, has garnered growing attention. Yet, despite their theoretical appeal, S4 models struggled with language modeling tasks, typically trailing Transformers by several points~\citep{ fu2022hungry, gu2021efficiently}.

Most recently, Mamba~\citep{gu2023mamba} advanced S4 models by enhancing their selectivity and effectiveness while enabling high training concurrency. It demonstrated strong performance on tasks involving long-range dependencies and compared favorably to Transformers in language modeling. However, training, scaling, and interpreting Mamba—and SSMs more broadly~\citep{smith2022simplified, poli2023hyena, hasani2022liquid}—remain challenging, while continue to be promising~\citep{dao2024transformers}.

We posit that the primary limitation of the Transformer lies in its inability to effectively model dependencies beyond its fixed context window. While its core self-attention mechanism is inherently parallelizable, this constraint makes its quadratic complexity a significant bottleneck {\em at scale}. This explains the surge of research aimed at reducing this complexity or exploring RNN-inspired alternatives. In this work, we propose a more viable approach by {\em decoupling} context width from sequence length, allowing models to scale to arbitrarily long sequences. Under this paradigm shift, the quadratic training complexity becomes less of a concern when small context windows are used, especially if the models maintain high parallelizability.

This paper introduces \textbf{Avey}\footnote{Avey is not an acronym, but a name that the authors like.}, a new architecture for language modeling that departs from Transformer-based and RNN-like designs. Avey is a flexible, sequence-length-invariant model that decouples sequence length from context width, thus enabling effective processing of long-range sequences. It preserves the influence of tokens that appear {\em outside} its context window, regardless of their positions in the sequence. This is achieved via a weighted-selective-split interaction mechanism, which systematically skips irrelevant tokens beyond the context window and ensures direct interactions with only relevant ones, retaining their contributions irrespective of sequence length.

\begin{figure}[t]
  \centering
  \includegraphics[width=\linewidth, height=8cm, keepaspectratio=true]{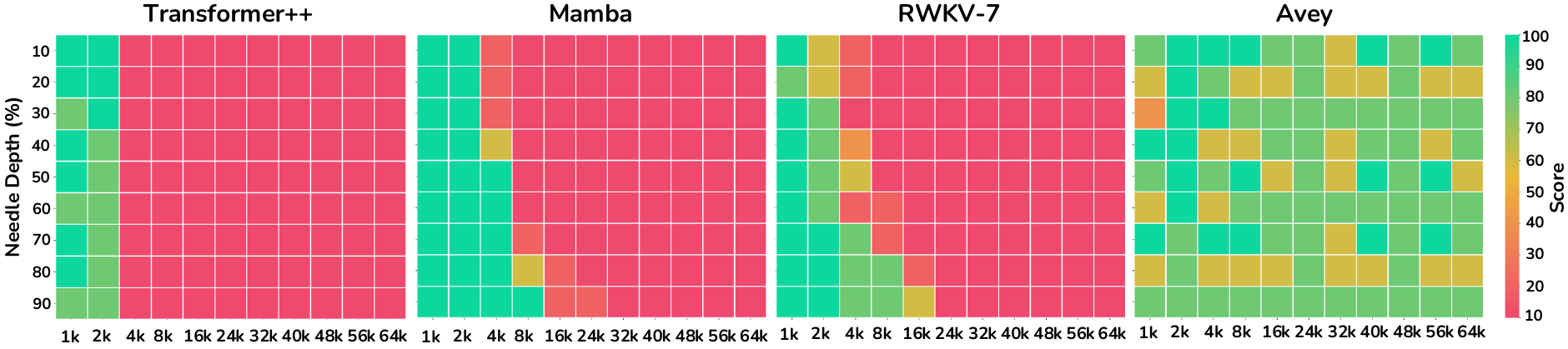}
  \caption{Needle-in-a-Haystack test performance comparison between Transformer++, Mamba, RWKV-7, and Avey, all using 1.5B parameters. The x-axis denotes the lengths of haystacks (i.e., documents with distractor texts, varying from 2k to 64k tokens) and the y-axis refers to the position of the needle (i.e., a short sentence) within each of the haystacks. A green cell indicates successful needle recall, while a red cell indicates failure. Transformer++, Mamba, and RWKV-7 were trained with 2k-token context windows, whereas Avey was trained with {\em only} a 512-token window yet was able to extrapolate to the longest sequences evaluated.}
  \label{fig:NiaH}
  \vspace{-1em}
\end{figure}

Fig.~\ref{fig:NiaH} illustrates Avey’s ability to generalize beyond its training context. A popular benchmark for evaluating this capability is Needle-in-a-Haystack (NiaH)~\citep{kamradt2023needle}. This benchmark measures a model’s capacity to recite a specific sentence (i.e., the {\em needle}) placed at an arbitrary position within a large body of distractor text (i.e., the {\em haystack}). Since its introduction, NiaH has become a widely used sandbox for probing the limits of long-context language models in capturing distant dependencies, and smaller models in generalizing beyond their trained context windows~\citep{fu2024data}. As shown in the figure, Transformer++ (i.e., the Transformer with an enhanced architecture and training recipe-- see Section~\ref{sec:experimental_methodology} for details), which was trained with a 2k-token context window, could not generalize beyond that limit. In contrast, Mamba and RWKV-7~\citep{peng2025rwkv}, also trained with 2k-token windows, managed to generalize to nearly 8k and 16k tokens, respectively. Most notably, Avey, despite being trained on a context window of {\em only} 512 tokens, successfully generalized to the maximum tested sequence length of 64k tokens, demonstrating strong extrapolative capability far beyond its original training regime.

To elaborate on its technical aspects, Avey is a recurrence- and attention-free architecture comprising two principal components, a ranker and a neural processor. The ranker slices each input sequence into splits of consecutive tokens and selects the top {\em k} most relevant splits for each current split being processed by the neural processor. The neural processor consists of three core units, the enricher, contextualizer, and fuser. The enricher enhances the quality of token embeddings by expanding their learnable features using a position-wise neural network. The contextualizer is an embedding-wise neural network with dynamic parameterization, enabling interactions between relevant tokens across the current and top {\em k} splits. Lastly, the fuser learns a function that integrates the contextualized features produced by the contextualizer with some uncontextualized features bypassed by a partial-embedding bypassing mechanism.

To summarize, our main contributions in this paper are as follows:
\begin{itemize}
  \item We propose Avey, a new recurrence- and attention-free neural architecture that decouples context window from sequence length, thus enabling effective processing of long-range sequences.
  
  \item We show that Avey performs comparably to the Transformer—outperforming it at two model sizes and underperforming it at one—across a range of popular zero-shot NLP benchmarks, thereby establishing an initial foundational architecture with potential for more scalable and effective language modeling.

  \item In contrast to the Transformer, we demonstrate that Avey can scale far beyond its context window using the standard Single Needle-In-A-Haystack (S-NIAH) benchmark suite from RULER~\citep{hsieh2024ruler}.

  \item We show that Mamba (representing SSMs) and RWKV-7 (representing linear attention models) exhibit some ability to generalize beyond their training context windows, but their performance decline significantly as the sequence length increases far beyond them. By comparison, Avey consistently and substantially outperforms both Mamba and RWKV-7 on the S-NIAH benchmark suite.
  
  \item We conduct extensive ablation studies to assess the impact of each design choice in Avey.

  \item We open-source the code and pretrained checkpoints of Avey to facilitate reproducibility and foster future research\footnote{\url{https://github.avey.ai/avey-dpa}.}.

\end{itemize}

\section{Avey}
\label{avey}

As indicated earlier, Avey comprises two components, a ranker and a neural processor. We next describe each component in detail (see Appendix~\ref{sec:design_rationale} for more design intuitions behind them).

\subsection{Ranker}
\label{sec:ranker}

\begin{figure}[t]
  \centering
  \includegraphics[width=\linewidth, height=5.5cm, keepaspectratio=true]{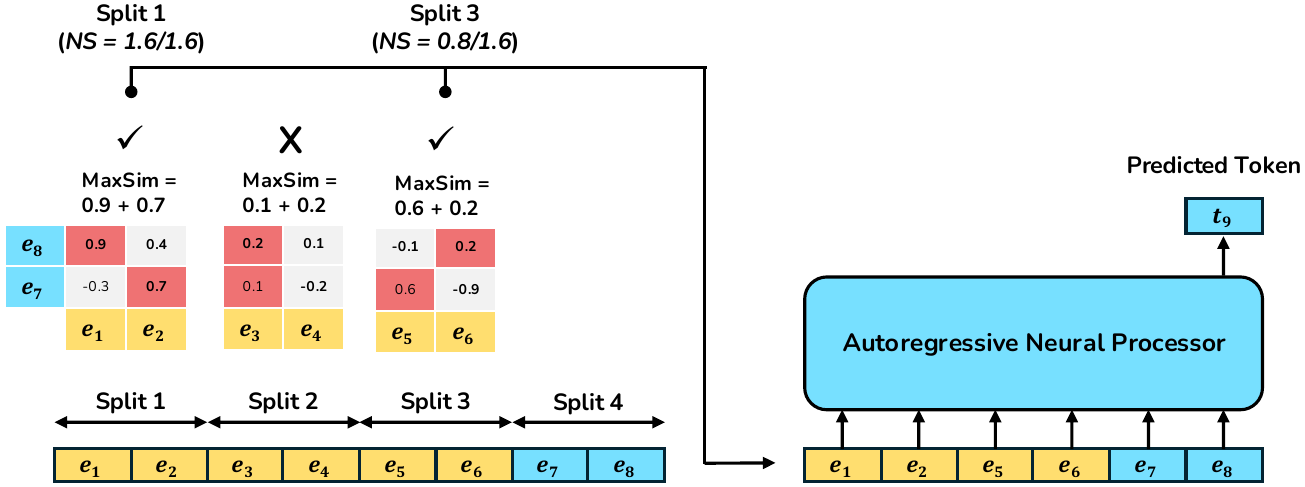}
  \caption{The ranker (left) partitions each input sequence into equal-sized splits and identifies the top $k$ most relevant ones (e.g., splits 1 and 3 for $k=2$) with respect to the \emph{current split} (e.g., split 4), using the MaxSim operator. These top-$k$ splits are then weighted by their normalized scores, where the normalized score (NS) of a split is computed as the ratio of its MaxSim value to the highest MaxSim score among the $k$ splits. Finally, the weighted top-$k$ splits are contextualized together with the current split by the neural processor (right).}
  \label{fig:ranker}
\end{figure}

Avey decouples sequence length from context width, enabling the processing of arbitrarily long sequences. The sequence length refers to the total number of tokens in a sequence, while the context width denotes the number of tokens that the neural processor can contextualize simultaneously. Importantly, the sequence length can be set to a value that is much larger than the context width. As such, the influence of {\em global} tokens (or tokens that fall outside the context window) may diminish as more tokens are successively processed. If such global tokens are semantically relevant to {\em local} tokens (or tokens that fall within the context window), the quality of token representations will decline, and the effectiveness of the model will degrade.

To this end, the ranker and neural processor jointly employ a \textbf{weighted-selective-split interaction mechanism}, which skips irrelevant global tokens and ensures direct interactions with relevant ones, preserving their impact regardless of sequence length. As demonstrated in Fig.~\ref{fig:ranker}, Avey divides each input sequence into equal-sized {\em splits}, each consisting of a list of contiguous token embeddings. Prior to predicting the next token in the sequence (e.g., token 9 in the figure), Avey involves the ranker to identify the top {\em k} (e.g., 2 in the figure) splits that are most relevant (e.g., splits 1 and 3) to the {\em current split} (i.e., split 4). 

The current split is defined as the one that either contains the token to be predicted (e.g., split 4 may contain only embedding 7, and Avey will aim to predict token 8) or contributes to predicting the {\em first} token in its subsequent split (e.g., split 4 serves in predicting token 9, which will belong to split 5 once predicted). To determine relevance, the ranker computes a similarity score between the current split, say $S_c$, and each preceding split, say $S_p$, using the MaxSim operator~\citep{khattab2020colbert}, originally proposed and utilized in Information Retrieval. Specifically, pairwise similarities are calculated (e.g., using a cosine function) between each embedding in $S_c$ and all embeddings in $S_p$. For each embedding in $S_c$, the maximum similarity across all $S_p$'s embeddings is taken, and then the maxima of all $S_c$'s embeddings are added to yield the final MaxSim score (see Fig.~\ref{fig:ranker}). This score signifies how relevant $S_p$ is to $S_c$.

Subsequently, the preceding splits are ranked based on their MaxSim scores, and the top {\em k} most relevant ones are contextualized with the current split by the neural processor, while maintaining their original order in the sequence. Before contextualization, however, the MaxSim scores of the top {\em k} splits are normalized with respect to the highest MaxSim score among them (e.g., split 3’s MaxSim score of 0.8 in Fig.~\ref{fig:ranker} is normalized via dividing it by the maximum score among the top {\em k} splits, i.e., 1.6, yielding 0.5). Each selected split is then {\em weighted} by its corresponding normalized MaxSim score, effectively scaling its contribution during contextualization.

As a result, the weighted-selective-split interaction mechanism does not only allow the ranker to rank splits based on relevance but also the neural processor to contextualize them accordingly, as each selected split is pre-weighted by its relevance score. This empowers the neural processor to judiciously leverage global information (i.e., splits beyond the context width) by focusing selectively on only relevant features, emphasizing informative ones and deemphasizing less useful ones, thus enhancing performance. We analyze the impact of weighting the top {\em k} splits in Section~\ref{sec:ablation}.

Note that the ranker is invoked {\em only once} per full forward and backward passes\footnote{It is worth emphasizing that the ranker is an \emph{internal} module that operates solely over \emph{in\mbox{-}sequence} splits already present in the input. In other words, it does not access external corpora or indexes, and therefore introduces neither retrieval latency nor corpus\mbox{-}freshness dependencies. Consequently, it is \emph{not} a RAG component, which retrieves \emph{external} evidence at inference (and/or training) time from a separate knowledge base. The two approaches are orthogonal in fact (the ranker allocates internal context, while RAG modifies the available evidence set). See Appendix~\ref{sec:ranker_and_rag} for a detailed discussion.}. To elucidate, Avey's depth can be increased by stacking multiple layers within the neural processor (see Fig.~\ref{fig:neural_processor}), thereby enabling the modeling of complex, hierarchical patterns. In contrast, only a single ranker is required before the stack of layers within the processor, regardless of their number. Once the ranker identifies the top {\em k} relevant splits of the current split, the neural processor contextualizes them all using one or more layers. 

Consequently, during training, each current split is matched once against every preceding split. This results in a compute cost of \( \frac{N/S \cdot (N/S + 1)}{2} \cdot S^2 d \) or a time complexity of \( O(N^2 d) \), where $N$ is the sequence length, $S$ is the split size, and $d$ is the embedding dimension. This complexity assumes that scalar multiply-add operations (e.g., those used in computing cosine similarity for MaxSim) and comparisons (e.g., those utilized to determine maximum scores) are constant-time.

\subsection{Neural Processor}
\label{neural_processor}

The neural processor  encompasses three key machineries, the {\em enricher}, {\em contextualizer}, and {\em fuser} (see Fig.~\ref{fig:neural_processor}). We describe each in detail below.

\subsubsection{The Enricher}
\label{enricher}

\begin{figure}[t]
  \centering
  \includegraphics[width=\linewidth, height=7cm, keepaspectratio]{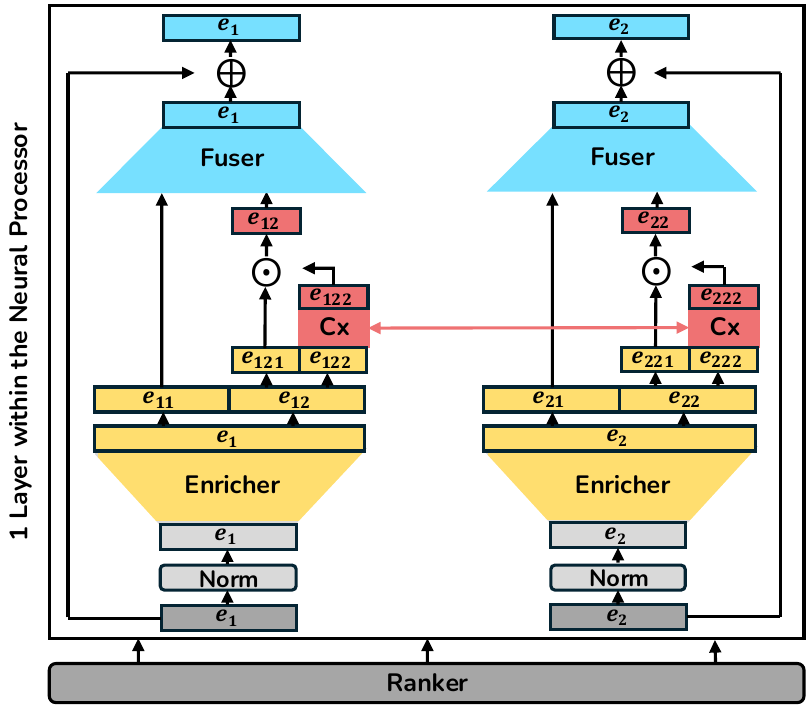}
  \caption{The neural processor (top) with its three major components, the enricher, contextualizer (Cx), and fuser. The processor is unfolded into two copies for illustrative purposes only, to show how different embeddings, (e.g., $e_1$ and $e_2$, or more precisely, parts of their tails, i.e., $e_{122}$ and $e_{222}$) are contextualized by Cx (i.e., in reality, all components are shared across all embeddings and many embeddings can be input to Cx simultaneously).}
  \label{fig:neural_processor}
  \vspace{-1em}
\end{figure}

The enricher aims at enriching the quality of each token representation via expanding the quantity of its learnable features, thereby enabling the contextualizer to capture more nuanced distinctions between tokens. Concretely, it is a one-layer, position-wise neural network (i.e., the input to each neuron is a single scalar element from an embedding), thus operating on each embedding independently, without considering neighboring embeddings. As such, it allows {\em intra-feature interactions} within the context of each individual embedding, facilitating the learning of higher-order and more expressive representations. The enricher can expand each input embedding by an arbitrary factor. We study the effect of varying the expansion factor on Avey’s performance in Section~\ref{sec:best_expansion_factor}, and ablate the enricher's contribution in Section~\ref{sec:ablation}.

Equation~\ref{eq:enricher} formalizes the enricher, where \( \mathbf{X} \in \mathbb{R}^{C \times d} \) is a matrix of \( C \) input embeddings (\( C \leq N \), where \( N \) is the sequence length), each of dimension \( d \); \( \sigma \) is an activation function; \( \mathbf{U} \in \mathbb{R}^{d \times m} \) is a learnable weight matrix defining a linear projection from dimension \( d \) to \( m \), where \( m > d \); and \( \mathbf{b} \in \mathbb{R}^{C \times m} \) denotes biases.

\begin{equation}
\mathbf{Z} = \sigma(\mathbf{X} \mathbf{U} + \mathbf{b})
\label{eq:enricher}
\end{equation}

As demonstrated in Fig.~\ref{fig:neural_processor}, the enricher feeds both, the contextualizer {\em and} the fuser. In particular, it bypasses a portion of each expanded embedding directly to the fuser in a technique that we refer to as \textbf{partial-embedding bypassing}. More formally, the output of the enricher, \( \mathbf{Z} \in \mathbb{R}^{C \times m} \), is split into two parts: (1) the {\em head} \( \mathbf{Z}_h \in \mathbb{R}^{C \times m_h} \), which is bypassed directly to the fuser, and (2) the {\em tail} \( \mathbf{Z}_t \in \mathbb{R}^{C \times m_t} \), which is forwarded to the contextualizer, where \( m = m_h + m_t \). Consequently, varying the tail size alters the head size, which can influence Avey’s performance. We investigate the impact of different tail sizes on Avey’s performance in Section~\ref{sec:best_tail_size}.

The partial-embedding bypassing technique allows preserving raw distinctive features of each embedding, thus inducing representations with inherent diversity. This diversity may serve in alleviating issues like entropy collapse~\citep{zhai2023stabilizing}, where the contextualizer increasingly focuses on a few tokens, and over-smoothing~\citep{zhou2021deepvit, shi2022revisiting, zhou2024value}, where embeddings become increasingly similar, as Avey’s depth is increased. We analyze the significance of partial-embedding bypassing on Avey’s effectiveness in Section~\ref{sec:ablation}.

Lastly, Equation~\ref{eq:enricher} implies that each neuron performs a weighted sum of \( d \) input features (i.e., the elements of an embedding), incurring \( d - 1 \) multiply-add operations. Since \( d \) is projected to a higher dimension \( m \)\footnote{In our case, we experiment with \( m \) being a multiple of \( d \), entailing that \( m \geq 2d \) (see Section~\ref{sec:best_expansion_factor}).}, the total computational cost is \( m(d - 1) \) per token. For a sequence of \( N \) tokens, the cost becomes \( Nm(d - 1) \), or asymptotically \( \mathcal{O}(Nmd) \).

\subsubsection{The Contextualizer}
\label{contextualizer}

The contextualizer is a one-layer, embedding-wise neural network (i.e., the input to each neuron is one embedding), thus operating in parallel on \( C \) embeddings, where \( C \) denotes the context width. More precisely, it enables inter-embedding, {\em data-dependent} interactions of only tail embeddings (i.e., \( \mathbf{Z}_t \in \mathbb{R}^{C \times m_t} \), as defined in Section~\ref{enricher}), after each enricher’s output embedding \( m \) is split into a head part (i.e., \( m_h \)) and a tail part (i.e., \( m_t \)), and only the \( m_t \) part (e.g., \( e_{12} \) and \( e_{22} \) in Fig.~\ref{fig:neural_processor}) is forwarded to the contextualizer.

The \( m_t \) part of each enriched embedding is further divided into two equal portions, \( m_{tl} \) (or left portion) and \( m_{tr} \) (or right portion), to enable judicious control of information flow through the neural processor. Specifically, \( m_{tl} \) serves as a {\em gating mechanism} for \( m_{tr} \), regulating how much of its contextualized feature values are propagated forward. Both \( m_{tl} \) and \( m_{tr} \) are learnable by the model, hence, allowing \( m_{tl} \) to dynamically capture the significance of each \( m_{tr} \)'s feature, and emphasize or deemphasize its influence accordingly. This gating mechanism was inspired from gMLP~\citep{liu2021pay} and resembles that of Gated Linear Units~\citep{dauphin2017language, shazeer2020glu, wu2019pay}.

More formally, \( \mathbf{Z}_t \in \mathbb{R}^{C \times m_t} \) is partitioned into two equal parts, \( \mathbf{Z}_{tl} \in \mathbb{R}^{C \times \left(m_t / 2\right)} \), which is bypassed to a multiplicative element-wise operation as part of a gating mechanism, and \( \mathbf{Z}_{tr} \in \mathbb{R}^{C \times \left(m_t / 2\right)} \), which is contextualized via a neural network, where each neuron takes as input an embedding of dimension \( m_t / 2 \). Equation~\ref{eq:contextualizer} defines the overall process, where \( \mathbf{V} \in \mathbb{R}^{C \times C} \) is a learnable weight matrix representing a linear cross-embedding transformation, \( \odot \) denotes element-wise multiplication, \( \mathbf{b}' \in \mathbb{R}^{C \times (m_t / 2)} \) refers to optional biases, and \( \mathcal{N}(\mathbf{Z}_{tr}) \) and \( \mathcal{N}(\mathbf{Z}_{tr}^\top) \) are row- and column-wise normalized versions of \( \mathbf{Z}_{tr} \), respectively.

\begin{equation}
\mathbf{c}(\mathbf{Z}_t) = \mathbf{Z}_{tl} \odot \sigma\left( \left( \mathbf{V} \odot \mathcal{N}(\mathbf{Z}_{tr}) \mathcal{N}(\mathbf{Z}_{tr}^\top) \right) \mathbf{Z}_{tr} + \mathbf{b}' \right)
\label{eq:contextualizer}
\end{equation}

Equation~\ref{eq:contextualizer} suggests that each neuron in the contextualizer’s network performs a weighted sum of the cosine similarities between embeddings (denoted by \( \mathcal{N}(\mathbf{Z}_{tr}) \mathcal{N}(\mathbf{Z}_{tr}^\top) \)) and the embeddings themselves (denoted by \( \mathbf{Z}_{tr} \)). This introduces a level of {\em selectivity} into the neural processor, as advocated by~\citep{gu2023mamba}. Specifically, it makes the parametrization of the neural processor dynamic, enabling it to disregard or focus on information during inference based on the input. We examine the influence of dynamic parametrization on Avey's performance in Section~\ref{sec:ablation}\footnote{See also a discussion on neural contextualization versus attention in Appendix~\ref{sec:contextualization_attention}.}. 

Finally, we note that the contextualizer inherently models the relationships between tokens, making the neural processor naturally aware of their positions in the sequence (i.e., positional encodings are not needed). In terms of complexity, as each neuron performs a weighted sum of \( C \) embeddings, each of dimension \( m_t / 2 \), it results in a cost of \( (C - 1) m_t / 2 \) multiply-add operations. With \( C \) neurons, the cost becomes \( C(C - 1) m_t / 2 \). For a sequence of \( N \) tokens, the contextualizer processes \( N / S \) splits, each contextualized with \( k \) relevant splits, making \( C = S(k + 1) \) and yielding a total cost of \( (N / S)[C(C - 1) m_t / 2] = N(k + 1)[(C - 1) m_t / 2] \) (after substituting \( S \) with \( C / (k + 1) \)), or asymptotically \( \mathcal{O}(N k C m_t) \).

\subsubsection{The Fuser}
\label{fuser}

The fuser is designed to learn an optimal function, referred to as {\em fusion}\footnote{The name is inspired from the CNN literature~\citep{hu2018squeeze}.}, that integrates uncontextualized features (i.e., those of dimension \( m_h \), bypassed by the partial-embedding bypassing technique) with contextualized features (i.e., those of dimension \( m_t / 2 \), output by the contextualizer). Subsequently, it produces, for each input token, a {\em contracted} representation that matches the token’s original embedding dimension \( d \) (see Fig.~\ref{fig:neural_processor}). Akin to the enricher, it is a one-layer, position-wise neural network, which operates on each embedding of dimension \( m_h + m_t / 2 \) independently.

Equation~\ref{eq:fuser} provides a mathematical definition of the fuser, where \( \mathbf{Z}_h \in \mathbb{R}^{C \times m_h} \) (as described in Section~\ref{enricher}) and \( \mathbf{c}(\mathbf{Z}_t) \in \mathbb{R}^{C \times (m_t / 2)} \) (as suggested by Equation~\ref{eq:contextualizer}) are concatenated, and \( \mathbf{O} \in \mathbb{R}^{(m_h + m_t / 2) \times d} \) is a learnable weight matrix representing a linear projection from dimension \( m_h + m_t / 2 \) back to dimension \( d \), where \( d < m_h + m_t / 2 \)\footnote{This inequality will always hold if \( m_h + m_t \geq 2d \), as is the case in our experiments (see Section~\ref{experiments}).}.

\begin{equation}
f(\mathbf{Z}) = [\mathbf{Z}_h \,\|\, \mathbf{c}(\mathbf{Z}_t)] \mathbf{O}
\label{eq:fuser}
\end{equation}

Equation~\ref{eq:fuser} entails that each neuron performs a weighted sum of \( m_h + m_t / 2 \) embedding elements, yielding a cost of \( (m_h + m_t / 2 - 1) \) multiply-add operations. For \( d \) neurons (since the fuser projects \( m_h + m_t / 2 \) to \( d \)), the cost is \( d(m_h + m_t / 2 - 1) \). For a sequence of \( N \) tokens, the total cost is \( N d (m_h + m_t / 2 - 1) \), or asymptotically \( \mathcal{O}(N m d) \).

Considering the aggregate computational costs of the ranker, enricher, contextualizer, and fuser, Avey exhibits a training time complexity of \( \mathcal{O}(L(2Nmd + NkC m_t) + N^2 d) \), where \( L \) denotes the number of neural processor layers. As the term \( N^2 d \) dominates asymptotically, the overall complexity simplifies to \( \mathcal{O}(N^2 d) \). During inference, the complexity reduces to \( \mathcal{O}(N) \), or linear per token. We provide a detailed analysis of Avey’s time complexity in Appendix~\ref{sec:complexity_analysis}.

\begin{figure}[t]
  \centering
  \includegraphics[width=\linewidth, keepaspectratio=true]{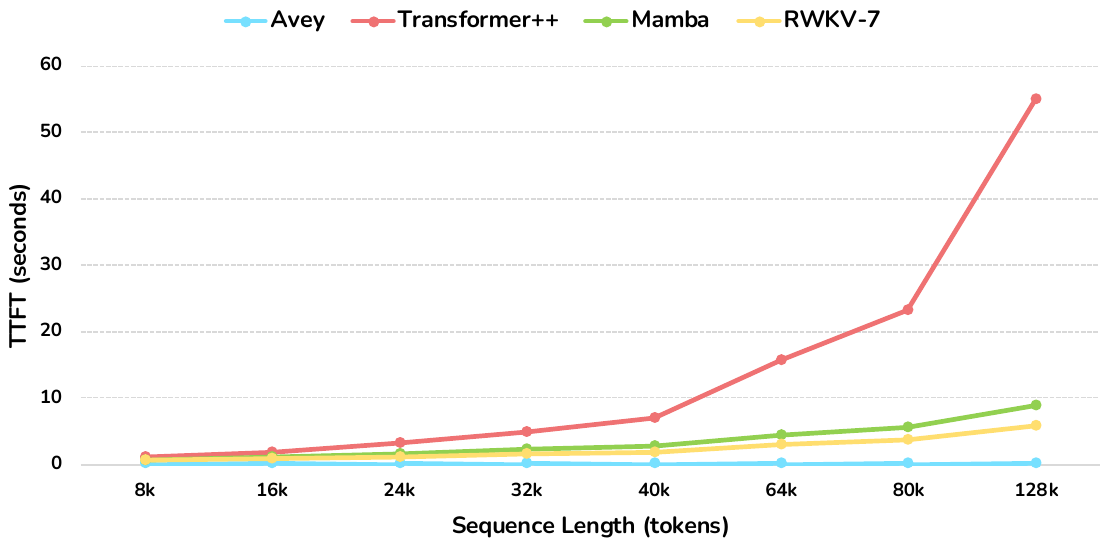}
  \caption{The Time to First Token (TTFT) for Avey, Transformer++, Mamba, and RWKV-7 across varying sequence lengths.}
  \label{fig:ttft}
\end{figure}

Lastly, to characterize Avey’s inference-time efficiency relative to other models, we benchmarked Time to First Token (TTFT)—a key latency metric for real-time applications~\citep{horton2024kv, liu2025speculative, dexter2025llm}—on a single NVIDIA H200 GPU. The evaluation included Avey, Transformer++, Mamba, and RWKV-7 across a range of sequence lengths. As shown in Fig.~\ref{fig:ttft}, Transformer++ demonstrates an approximately quadratic increase in TTFT as the sequence length \( N \) grows, owing to its full self-attention mechanism, which must operate over the entire prompt before generating the first token. In contrast, Mamba and RWKV-7 scale linearly with \( N \), as they require a full forward pass to construct their RNN-style hidden states prior to emitting the first token. While Avey is also theoretically expected to scale linearly, its empirical TTFT is significantly lower than that of Transformer++, Mamba, and RWKV-7. This stems from Avey’s architectural design, which invokes its dominant contributor to inference complexity (see Appendix~\ref{sec:complexity_analysis}), namely, the ranker, {\em only once} per forward pass. As a result, the ranker imposes minimal computational overhead in practice, enabling Avey to achieve substantially lower TTFT and making it particularly well-suited for latency-sensitive, real-world applications (e.g., conversational AI and edge deployments).

\section{Experiments}
\label{experiments}

\subsection{Methodology}
\label{sec:experimental_methodology}

We now describe our experimental methodology in detail. To begin with, we adopt a \emph{cascaded search} to identify the best configuration of Avey. Starting from a baseline neural processor ({\em excluding} the ranker), we sequentially evaluate individual architectural choices. After each empirical finding, we integrate the best-performing architectural element into the processor (one at a time) and resume the search process from the updated configuration. This cascaded procedure unfolds chronologically in Sections~\ref{sec:activation_enricher}, \ref{sec:activation_contextualizer}, \ref{sec:best_expansion_factor}, \ref{sec:best_tail_size}, and \ref{sec:embedding_dim}.

More precisely, we initiate the cascaded search with an expansion factor of \(4\times\) in the enricher, a tail size of \(50\%\) (i.e., half of each expanded embedding is forwarded to the contextualizer), RMSNorm~\citep{zhang2019rmsnorm} for normalization, no activations in either the enricher or contextualizer, a global batch size of 0.5M, a context width of 1024, and a constant learning rate of $1e{-3}$. As each architectural element is explored and empirically selected, we incorporate it into the processor and resume the search from the updated configuration. For instance, after identifying \(\mathrm{ReLU}^2\) as the most effective activation for the enricher, we included it in the architecture and proceeded with the remaining explorations.

After completing the exploratory experiments described above, we integrated the ranker into the neural processor and conducted an extensive study—spanning over 138 training and inference runs—to identify the optimal sequence length (\(N\)), split size (\(S\)), and number of top-\(k\) splits (\(k\)). The results of this sensitivity analysis are summarized in Section~\ref{sec:sensitivity_ranker}. Subsequently, we evaluated normalization strategies in Section~\ref{sec:norm_study} and tuned the peak learning rate and learning-rate schedule for the full architecture in Section~\ref{sec:lr_schedules}.

\begin{table*}[t]
\centering
\caption{Training and model hyperparameters for Avey, Transformer++, Mamba, and RWKV-7 at three scales.}
\label{models_params}
\setlength{\tabcolsep}{3pt}
\renewcommand{\arraystretch}{1.15}
\begin{adjustbox}{max width=\textwidth}
\begin{tabular}{lccc|ccc|ccc|ccc}
\hline
 & \multicolumn{3}{c|}{Avey} & \multicolumn{3}{c|}{\textbf{Transformer++}} & \multicolumn{3}{c|}{\textbf{Mamba}} & \multicolumn{3}{c}{\textbf{RWKV-7}} \\
\hline
\textbf{Scale} & Small & Medium & Large & Small & Medium & Large & Small & Medium & Large & Small & Medium & Large \\
\hline
\textbf{\# of Parameters} & 153M & 496M & 1.52B & 152M & 488M & 1.5B & 144M & 500M & 1.4B & 168M & 501M & 1.5B \\
\textbf{Optimizer} & \multicolumn{3}{c|}{AdamW} & \multicolumn{3}{c|}{AdamW} & \multicolumn{3}{c|}{AdamW} & \multicolumn{3}{c}{AdamW} \\
\textbf{Betas} & \multicolumn{3}{c|}{(0.9, 0.95)} & \multicolumn{3}{c|}{(0.9, 0.95)} & \multicolumn{3}{c|}{(0.9, 0.95)} & \multicolumn{3}{c}{(0.9, 0.95)} \\
\textbf{Epsilon} & \multicolumn{3}{c|}{$1\mathrm{e}{-12}$} & \multicolumn{3}{c|}{$1\mathrm{e}{-12}$} & \multicolumn{3}{c|}{$1\mathrm{e}{-12}$} & \multicolumn{3}{c}{$1\mathrm{e}{-12}$} \\
\textbf{Peak learning rate} &
$1\mathrm{e}{-3}$ & $3\mathrm{e}{-3}$ & $1.5\mathrm{e}{-3}$ &
$1.25\mathrm{e}{-3}$ & $3\mathrm{e}{-3}$ & $1.5\mathrm{e}{-3}$ &
$1.0\mathrm{e}{-3}$ & $6\mathrm{e}{-4}$ & $4\mathrm{e}{-4}$ &
$6\text{e}{-4}$ & $4\text{e}{-4}$ & $4\text{e}{-4}$ \\
\textbf{Schedule} &
\multicolumn{3}{c|}{\makecell{Cosine decay to 10\% of peak;\\ no warmup}} &
\multicolumn{3}{c|}{\makecell{Linear warmup 10\% of steps;\\ cosine decay to 10\%}} &
\multicolumn{3}{c|}{\makecell{Linear warmup 10\% of steps;\\ cosine decay to 10\%}} &
\multicolumn{3}{c}{\makecell{Cosine decay to 10\% of peak;\\ no warmup}} \\
\textbf{Batch size} & 0.5M & 0.5M & 1M & 0.5M & 0.5M & 1M & 0.5M & 0.5M & 1M & 1M & 1M & 2M \\
\textbf{Gradient norm clip} & \multicolumn{3}{c|}{1.0} & \multicolumn{3}{c|}{1.0} & \multicolumn{3}{c|}{1.0} & \multicolumn{3}{c}{1.0} \\
\textbf{Weight decay} & \multicolumn{3}{c|}{0.1 (matrices only)} & \multicolumn{3}{c|}{0.1 (matrices only)} & \multicolumn{3}{c|}{0.1 (matrices only)} & \multicolumn{3}{c}{0.1 (matrices only)} \\
\textbf{Context width} & \multicolumn{3}{c|}{512} & \multicolumn{3}{c|}{2048} & \multicolumn{3}{c|}{2048} & \multicolumn{3}{c}{2048} \\
\textbf{Split size ($S$)} & \multicolumn{3}{c|}{64} & \multicolumn{3}{c|}{N/A} & \multicolumn{3}{c|}{N/A} & \multicolumn{3}{c}{N/A} \\
\textbf{Top-$k$ splits} & \multicolumn{3}{c|}{7} & \multicolumn{3}{c|}{N/A} & \multicolumn{3}{c|}{N/A} & \multicolumn{3}{c}{N/A} \\
\textbf{Vocabulary size} & \multicolumn{3}{c|}{50{,}304} & \multicolumn{3}{c|}{50{,}304} & \multicolumn{3}{c|}{50{,}304} & \multicolumn{3}{c}{50{,}304} \\
\textbf{Expansion factor} & \multicolumn{3}{c|}{4} & \multicolumn{3}{c|}{N/A} & \multicolumn{3}{c|}{N/A} & \multicolumn{3}{c}{N/A} \\
\textbf{Tail size} & \multicolumn{3}{c|}{0.5} & \multicolumn{3}{c|}{N/A} & \multicolumn{3}{c|}{N/A} & \multicolumn{3}{c}{N/A} \\
\textbf{Embedding dimension} & 768 & 768 & 2048 & 768 & 1024 & 1664 & 768 & 1280 & 2048 & 768 & 1024 & 2048 \\
\textbf{Number of layers} & 26 & 104 & 48 & 12 & 26 & 32 & 28 & 42 & 52 & 12 & 30 & 24 \\
\textbf{Number of heads} & \multicolumn{3}{c|}{N/A} & \multicolumn{3}{c|}{12} & \multicolumn{3}{c|}{N/A} & \multicolumn{3}{c}{16} \\
\hline
\end{tabular}
\end{adjustbox}
\end{table*}

All the aforementioned experiments were conducted with a 145M-parameter model trained on 10B tokens from the FineWeb dataset\footnote{This dataset is released under the Open Data Commons Attribution License (ODC-By) v1.0.}~\citep{fineweb2023} (specifically, the \texttt{sample-100BT} subset). These results informed the final choices of training and model hyperparameters for Avey across three parameter scales, 153M (\emph{small}), 496M (\emph{medium}), and 1.52B (\emph{large}), as reported in Table~\ref{models_params}.

For baselines, we compare Avey against three leading open-source models, namely, Transformer++~\citep{nanogpt2023}, Mamba~\citep{mambai2023}, and RWKV-7~\citep{rwkvi2023}. For Transformer++, we implement the strongest recipe known to us, incorporating rotary positional encodings (RoPE)~\citep{su2024roformer}, SwiGLU MLPs~\citep{shazeer2020glu}, and RMSNorm in place of LayerNorm~\citep{zhang2019rmsnorm}. All models are trained with their best-known hyperparameters (see Table~\ref{models_params}) at three parameter scales (i.e., \emph{small}, \emph{medium}, and \emph{large}) under a fixed budget of 100B tokens drawn from the FineWeb dataset\footnote{More precisely, all models---Avey, Transformer++, Mamba, and RWKV-7---are trained for one epoch over the \texttt{sample-100BT} subset of FineWeb.}. For consistency and comparability, we use the \texttt{p50k\_base} tokenizer~\citep{tiktoken2022} across all models, matching the GPT-2--derived token counts reported for this dataset.

To compare all models, we employ a suite of widely used NLP benchmarks, including ARC-E and ARC-C (for scientific reasoning and reading comprehension)~\citep{clark2018think}, HellaSwag (for commonsense inference)~\citep{zellers2019hellaswag}, PIQA (for physical reasoning)~\citep{bisk2020piqa}, OBQA (for open-book science reasoning)~\citep{mihaylov2018can}, SIQA (for social interaction understanding)~\citep{sap2019socialiqa}, and Winogrande (for coreference and commonsense reasoning)~\citep{sakaguchi2021winogrande}. In addition, we evaluate long-context retrieval capabilities using the standard Single Needle-In-A-Haystack (S-NIAH) benchmark suite from RULER~\citep{hsieh2024ruler}, which measures a model’s ability to recover passkeys from large distractor contexts at sequence lengths ranging from 2k to 64k tokens. All evaluations are conducted with the widely adopted LM Evaluation Harness from EleutherAI~\citep{gao2021framework}, consistent with prior work~\citep{gu2023mamba}.

For all models, we report performance in terms of benchmark accuracy\footnote{Additionally, all reported perplexity values in this paper refer to \emph{training} perplexity.}. When available, we use the normalized-accuracy metric (\texttt{acc\_norm}) from the LM Evaluation Harness. To mitigate variance due to training stochasticity, each model’s score on a given NLP benchmark is computed as the mean accuracy over its final three checkpoints (at 90B, 95B, and 100B tokens). Complete results across these checkpoints, along with summary statistics and discussion, are provided in Appendix~\ref{sec:extra_short_range_results}.

\begin{table*}[t]
\centering
\small
\caption{Summary of studies for key design choices and corresponding experimental references.}
\label{ablation_summary}
\setlength{\tabcolsep}{4pt}  
\resizebox{\textwidth}{!}{%
\begin{tabular}{>{\raggedright\arraybackslash}p{6.3cm} >{\raggedright\arraybackslash}p{6.2cm} >{\raggedright\arraybackslash}p{3.0cm}}
\toprule
\textbf{Question} & \textbf{Answer} & \textbf{Experiments} \\
\midrule

Activation in the enricher? & Yes, ReLU\(^2\) & Section~\ref{sec:activation_enricher} \\

\cmidrule(lr){1-3}
Activation in the contextualizer? & No & Section~\ref{sec:activation_contextualizer} \\

\cmidrule(lr){1-3}
Enrich embeddings before contextualization? & Yes, by 4x & Sections~\ref{sec:ablation}, \ref{sec:best_expansion_factor} \\

\cmidrule(lr){1-3}
Bypass uncontextualized features to the fuser? & Yes, 50\% of each enriched embedding & Sections~\ref{sec:ablation}, \ref{sec:best_tail_size} \\

\cmidrule(lr){1-3}
Deeper model and narrower embeddings, or shallower model and wider embeddings? & Deeper model and narrower embeddings & Section~\ref{sec:embedding_dim} \\

\cmidrule(lr){1-3}
Best values for sequence length \(N\), split size \(S\), and top-\(k\) splits? & \(N = 512\), \(S = 64\), \(k = 7\) & Section~\ref{sec:sensitivity_ranker} \\

\cmidrule(lr){1-3}

RMSNorm or LayerNorm? & RMSNorm & Section~\ref{sec:norm_study} \\
\cmidrule(lr){1-3}
LR: to decay or not to decay? & Yes, cosine decay with peak LR of \(1\text{e}{-3}\) & Section~\ref{sec:lr_schedules} \\

\cmidrule(lr){1-3}
Weight ranked splits? & Yes, using normalized scores & Section~\ref{sec:ablation} \\

\cmidrule(lr){1-3}
Static or dynamic parametrization for the contextualizer? & Dynamic parametrization & Section~\ref{sec:ablation} \\
\cmidrule(lr){1-3}
Replace the contextualizer with self-attention? & No & Section~\ref{sec:ablation} \\
\bottomrule
\end{tabular}}
\end{table*}

Finally, all training and evaluation runs were executed on 208 NVIDIA H200 GPUs with mixed precision (bfloat16) enabled for training. Across the three model sizes and a 100B-token budget, the total wall-clock training time was approximately 80--90 hours under near-ideal parallelization using the above referenced implementations. To minimize sources of randomness and ensure comparability, we disabled \texttt{torch.compile} during all design-choice and sensitivity experiments. For ablations and final training runs, we enabled \texttt{torch.compile} when compatible to accelerate training. We also fixed the random seed to 11 (arbitrarily chosen) for all runs to further reduce variability due to stochastic effects.

\subsection{Design Choices}
\label{sec:design_choices}

We conducted more than 200 experiments to investigate key design choices. Table~\ref{ablation_summary} summarizes the main findings and cites the corresponding experiments that support each conclusion. We next elaborate on all these experiments.

\subsubsection{Activation or No Activation in the Enricher}
\label{sec:activation_enricher}

\begin{table*}[ht]
\centering
\small
\caption{Avey’s performance \emph{without} and \emph{with} an activation function in the \textbf{enricher}. The study isolates the neural processor (excluding the ranker) and trains it on 10B tokens.}
\label{tab:activation_comparison_enricher}
\setlength{\tabcolsep}{4pt}  
\resizebox{\textwidth}{!}{%
\begin{tabular}{lccccccccc}
\toprule
\textbf{Configuration} & \textbf{Perplexity} & \textbf{ARC-C} & \textbf{ARC-E} & \textbf{HellaSwag} &
\textbf{OBQA} & \textbf{PIQA} & \textbf{SIQA} & \textbf{Winogrande} & \textbf{Average} \\
\midrule
No Activation & 37.65 & 22.53 & 35.19 & 28.95 & 27.40 & 60.45 & 36.13 & 48.70 & 37.05 \\
GELU          & \textbf{30.10} & 23.04 & 37.88 & 31.37 & 26.00 & 63.55 & 36.95 & 51.85 & 38.66 \\
ReLU          & 31.02 & 22.87 & 37.21 & 31.35 & 28.00 & 62.79 & 38.13 & 48.86 & 38.46 \\
ReLU$^{2}$    & 30.18 & 24.15 & 38.76 & 32.08 & 28.00 & 63.76 & 38.28 & 50.12 & \textbf{39.31} \\
SiLU          & 30.81 & 22.18 & 38.43 & 31.30 & 28.20 & 62.30 & 37.05 & 53.20 & 38.95 \\
\bottomrule
\end{tabular}}
\end{table*}

In this study, we evaluate Avey’s performance \emph{with} and \emph{without} an activation function in the enricher. We train a 145M-parameter model on 10B tokens from the FineWeb dataset~\citep{fineweb2023}. The configuration uses an enricher expansion factor of \(4\times\) (each embedding is expanded fourfold), a tail size of \(50\%\) (i.e., half of each expanded embedding is forwarded to the contextualizer), RMSNorm~\citep{zhang2019rmsnorm} for normalization, and no activation in the contextualizer. The context width (i.e., the maximum number of tokens processed concurrently by the contextualizer) is set to 1024, and we employ a constant learning rate of $1e{-3}$ throughout training.

This study excludes the ranker and focuses solely on the neural processor. Besides, it evaluates four activation functions in the enricher, namely, GELU~\citep{hendrycks2016gelu}, ReLU~\citep{nair2010relu}, \(\mathrm{ReLU}^2\)~\citep{chowdhery2022palm}, and SiLU~\citep{ramachandran2017swish}. All other experimental settings follow the methodology in Section~\ref{sec:experimental_methodology}. Results are summarized in Table~\ref{tab:activation_comparison_enricher}. As shown, \(\mathrm{ReLU}^2\) yields a clear accuracy improvement over the no-activation baseline and is therefore adopted as the default enricher activation for Avey. Notably, GELU attains the lowest perplexity, though the gap relative to \(\mathrm{ReLU}^2\) is minimal. While perplexity measures next-token predictive quality on the training distribution, it is an indirect proxy for modeling capability and does not always \emph{precisely} predict downstream task performance.

\subsubsection{Activation or No Activation in the Contextualizer}
\label{sec:activation_contextualizer}

\begin{table*}[ht]
\centering
\small
\caption{Avey’s performance \emph{without} and \emph{with} an activation function in the \textbf{contextualizer}. The study isolates the neural processor (without the ranker), trains on 10B tokens, and fixes \(\mathrm{ReLU}^2\) in the enricher, following the findings in Table~\ref{tab:activation_comparison_enricher}.}
\label{tab:activation_comparison_contextualizer}
\setlength{\tabcolsep}{4pt}           
\resizebox{\textwidth}{!}{%
\begin{tabular}{lccccccccc}
\toprule
\textbf{Configuration} & \textbf{Perplexity} & \textbf{ARC-C} & \textbf{ARC-E} & \textbf{HellaSwag} &
\textbf{OBQA} & \textbf{PIQA} & \textbf{SIQA} & \textbf{Winogrande} & \textbf{Average} \\
\midrule
No Activation & \textbf{30.18} & 24.15 & 38.76 & 32.08 & 28.00 & 63.76 & 38.28 & 50.12 & \textbf{39.31} \\
GELU          & 30.64 & 22.01 & 37.54 & 31.23 & 27.20 & 64.74 & 37.21 & 50.67 & 38.66 \\
ReLU          & 30.30 & 23.29 & 37.84 & 31.82 & 26.60 & 64.15 & 38.13 & 50.59 & 38.92 \\
ReLU$^{2}$    & 31.05 & 22.35 & 38.80 & 30.78 & 27.20 & 63.49 & 37.31 & 52.01 & 38.85 \\
SiLU          & 30.92 & 23.63 & 36.74 & 31.51 & 27.40 & 64.20 & 36.18 & 50.04 & 38.53 \\
\bottomrule
\end{tabular}}
\end{table*}

Next, we evaluate Avey both \emph{with} and \emph{without} an activation function in the contextualizer. 
We employ the same experimental setup as in Section~\ref{sec:activation_enricher} and, consistent with the findings reported therein, fix \(\mathrm{ReLU}^2\) as the activation for the enricher. 
We then assess four candidate activations for the contextualizer, namely, GELU, ReLU, \(\mathrm{ReLU}^2\), and SiLU. 
The results, summarized in Table~\ref{tab:activation_comparison_contextualizer}, indicate that the best performance is achieved \emph{without} any activation in the contextualizer. 
Accordingly, we adopt this configuration as Avey’s default setting.

\subsubsection{What is the Best Expansion Factor?}
\label{sec:best_expansion_factor}

\begin{table*}[ht]
\centering
\small
\caption{The effect of the \emph{expansion factor} on Avey's performance. The study isolates the neural processor (without the ranker), trains on 10B tokens, fixes \(\mathrm{ReLU}^2\) as the enricher activation, and uses no activation in the contextualizer, following the configurations established in Tables~\ref{tab:activation_comparison_enricher} and~\ref{tab:activation_comparison_contextualizer}.}
\label{tab:expansion_factor}
\setlength{\tabcolsep}{4pt}            
\resizebox{\textwidth}{!}{%
\begin{tabular}{lccccccccc}
\toprule
\textbf{Expansion} & \textbf{Perplexity} & \textbf{ARC-C} & \textbf{ARC-E} & \textbf{HellaSwag} &
\textbf{OBQA} & \textbf{PIQA} & \textbf{SIQA} & \textbf{Winogrande} & \textbf{Average} \\
\midrule
2× & 30.85 & 23.29 & 36.83 & 30.92 & 26.20 & 63.98 & 37.72 & 51.78 & 38.67 \\
4× & 30.18 & 24.15 & 38.76 & 32.08 & 28.00 & 63.76 & 38.28 & 50.12 & \textbf{39.31} \\
8× & \textbf{30.00} & 23.21 & 37.79 & 31.48 & 26.20 & 63.82 & 36.59 & 50.75 & 38.55 \\
\bottomrule
\end{tabular}}
\end{table*}

In this study, we investigate the effect of the \emph{expansion factor} in the enricher, defined as the degree to which each input embedding is expanded. Specifically, we evaluate expansion factors ranging from \(2\times\) to \(8\times\) (see Table~\ref{tab:expansion_factor}) while maintaining a fixed total parameter budget (e.g., \(2\times\) uses 34 layers, whereas \(4\times\) uses 20 layers). The experimental setup follows that of Section~\ref{sec:activation_enricher}, employing \(\mathrm{ReLU}^2\) in the enricher~\citep{chowdhery2022palm} and no activation in the contextualizer, consistent with the findings from Sections~\ref{sec:activation_enricher} and~\ref{sec:activation_contextualizer}. As shown in Table~\ref{tab:expansion_factor}, an expansion factor of \(4\times\) delivers the strongest performance and is therefore adopted as the default configuration for the enricher.

\subsubsection{What is the Best Tail Size?}
\label{sec:best_tail_size}

\begin{table*}[ht]
\centering
\small
\caption{The effect of the \emph{tail size} on Avey's performance. The study isolates the neural processor (without the ranker), trains on 10B tokens, fixes \(\mathrm{ReLU}^2\) as the enricher activation, uses no activation in the contextualizer, and adopts an expansion factor of \(4\times\), following the findings in Tables~\ref{tab:activation_comparison_enricher}, \ref{tab:activation_comparison_contextualizer}, and \ref{tab:expansion_factor}.}
\label{tab:tail_size}
\setlength{\tabcolsep}{4pt}            
\resizebox{\textwidth}{!}{%
\begin{tabular}{lccccccccc}
\toprule
\textbf{Tail Size} & \textbf{Perplexity} & \textbf{ARC-C} & \textbf{ARC-E} & \textbf{HellaSwag} &
\textbf{OBQA} & \textbf{PIQA} & \textbf{SIQA} & \textbf{Winogrande} & \textbf{Average} \\
\midrule
10\% & 34.23 & 21.59 & 36.78 & 30.34 & 27.0 & 63.60 & 37.15 & 50.28 & 38.11 \\
30\% & 30.91 & 23.55 & 37.33 & 31.41 & 29.4 & 64.25 & 37.15 & 51.22 & 39.19 \\
50\% & 30.18 & 24.15 & 38.76 & 32.08 & 28.0 & 63.76 & 38.28 & 50.12 & \textbf{39.31} \\
70\% & \textbf{29.79} & 23.04 & 38.26 & 31.68 & 28.4 & 63.55 & 37.36 & 49.96 & 38.89 \\
90\% & 30.20 & 23.04 & 38.38 & 32.13 & 27.6 & 64.04 & 37.67 & 50.91 & 39.11 \\
\bottomrule
\end{tabular}}
\end{table*}

We now examine the impact of forwarding a \emph{tail} portion of each enriched embedding to the contextualizer. Concretely, we vary the \emph{tail size} (or the fraction of each enriched embedding passed to the contextualizer) from \(10\%\) to \(90\%\) (see Table~\ref{tab:tail_size}). The study follows the setup outlined in Section~\ref{sec:activation_enricher}, using \(\mathrm{ReLU}^2\) in the enricher, no activation in the contextualizer, and an expansion factor of \(4\times\), consistent with the findings in Sections~\ref{sec:activation_enricher}, \ref{sec:activation_contextualizer}, and~\ref{sec:best_expansion_factor}. As depicted in Table~\ref{tab:tail_size}, a tail size of \(50\%\) yields the strongest performance and is accordingly employed as Avey’s default configuration.

\subsubsection{Deeper Models and Narrower Embeddings, or Shallower Models and Wider Embeddings}
\label{sec:embedding_dim}

\begin{table*}[ht]
\centering
\small
\caption{Avey’s performance across configurations that trade embedding width for depth, including wider embeddings with fewer layers (e.g., 1536-dimensional embeddings with 24 layers for the 0.5B-parameter model) versus narrower embeddings with more layers (e.g., 768-dimensional embeddings with 90 layers for the 0.5B-parameter model). Throughout the text, the 140M-, 0.5B-, and 1.5B-parameter models are referred to as \emph{small Avey}, \emph{medium Avey}, and \emph{large Avey}, respectively.}
\label{tab:depth_vs_width}
\setlength{\tabcolsep}{4pt}            
\resizebox{\textwidth}{!}{%
\begin{tabular}{lccccccccccc}
\toprule
\textbf{\# Params} & \textbf{Embed.} & \textbf{\# Layers} &
\textbf{Perplexity} & \textbf{ARC-C} & \textbf{ARC-E} & \textbf{HellaSwag} &
\textbf{OBQA} & \textbf{PIQA} & \textbf{SIQA} & \textbf{Winogrande} &
\textbf{Average} \\
\midrule
\multirow{3}{*}{140 M}
& 512  & 40 & 31.14 & 22.53 & 37.58 & 31.05 & 28.6 & 63.87 & 37.46 & 52.64 & 39.10 \\
& 768  & 20 & \textbf{30.18} & 24.15 & 38.76 & 32.08 & 28.0 & 63.76 & 38.28 & 50.12 & \textbf{39.31} \\
& 1024 & 11 & 31.46 & 23.46 & 38.30 & 30.78 & 27.0 & 63.93 & 37.72 & 48.86 & 38.58 \\
\midrule
\multirow{3}{*}{0.5 B}
& 768  & 90 & \textbf{23.02} & 23.55 & 42.05 & 38.61 & 30.2 & 66.10 & 39.20 & 51.78 & 41.64 \\
& 1024 & 54 & 23.27 & 24.40 & 41.92 & 38.46 & 29.2 & 67.19 & 38.74 & 51.46 & 41.62 \\
& 1536 & 24 & 23.51 & 23.98 & 42.85 & 37.79 & 29.4 & 66.97 & 38.89 & 51.78 & \textbf{41.67} \\
\midrule
\multirow{3}{*}{1.5 B}
& 1536 & 80 & 19.97 & 25.26 & 45.50 & 44.56 & 30.2 & 68.61 & 40.02 & 52.33 & 43.78 \\
& 2048 & 48 & \textbf{19.84} & 25.77 & 46.55 & 44.99 & 31.6 & 69.42 & 40.17 & 52.17 & \textbf{44.38} \\
& 2560 & 30 & 20.23 & 26.62 & 45.16 & 43.91 & 29.2 & 69.10 & 39.82 & 52.09 & 43.70 \\
\bottomrule
\end{tabular}}
\end{table*}

The objective of this study is to determine whether a narrower embedding dimension with a greater model depth yields better or worse performance than a wider embedding dimension with fewer layers. The study utilizes the experimental setup described in Section~\ref{sec:activation_enricher} and leverages the findings of Sections~\ref{sec:activation_enricher},~\ref{sec:activation_contextualizer},~\ref{sec:best_expansion_factor}, and~\ref{sec:best_tail_size}. Consequently, it employs \(\mathrm{ReLU}^2\) as the activation function in the enricher, no activation in the contextualizer, an expansion factor of \(4\times\), and a tail size of \(50\%\).

We begin by evaluating a 140M-parameter Avey model (referred to as \emph{small Avey}) with three embedding dimensions, 512, 768, and 1024. To maintain a constant parameter budget, these dimensions yielded 40, 20, and 11 layers, respectively. Results are summarized in Table~\ref{tab:depth_vs_width}. As shown, the configuration with a 768-dimensional embedding and 20 layers outperforms the other two.

We next evaluate a 500M-parameter Avey model (denoted as \emph{medium Avey}) with embedding dimensions of 768, 1024, and 1536. To keep the total parameter count fixed, these settings resulted in 90, 54, and 24 layers, respectively. As reported in Table~\ref{tab:depth_vs_width}, the configuration with a 768-dimensional embedding and 90 layers achieves the strongest performance among the three.

Lastly, we examine a 1.5B-parameter Avey model (referred to as \emph{large Avey}) with embedding dimensions of 1536, 2048, and 2560. Again, to maintain a constant parameter budget, these settings yielded 80, 48, and 30 layers, respectively. As illustrated in Table~\ref{tab:depth_vs_width}, the configuration with a 2048-dimensional embedding and 48 layers delivers the best performance among the three.

The above results suggest a trend, whereby wider embedding dimensions (e.g., 1024 in small Avey; 1563 in medium Avey; and 2560 in large Avey) paired with shallower architectures (e.g., 11 layers in small Avey; 24 layers in medium Avey; and 30 layers in large Avey) tend to underperform deeper models (e.g., 40 and 20 layers in small Avey; 90 and 54 layers in medium Avey; and 80 and 48 layers in large Avey) with narrower embeddings (e.g., 512 and 768 in small Avey; 768 and 1024 in medium Avey; and 1536 and 2048 in large Avey). As such, in all our subsequent experiments, we adopt deeper models with narrower embedding dimensions, namely, the best performing small Avey, medium Avey, and large Avey in Table~\ref{tab:depth_vs_width}.

Interestingly, Table~\ref{tab:depth_vs_width} indicates that some benchmarks benefit more from increased model capacity than others. For example, the commonsense reasoning task HellaSwag shows gains of \(20.36\%\) and \(28.5\%\) under the 0.5B- and 1.5B-parameter models, respectively, relative to the 140M-parameter baseline. In contrast, the question–answering benchmark SIQA exhibits only a modest improvement of \(2.4\%\) for \emph{both} the 0.5B- and 1.5B-parameter models, suggesting lower sensitivity to model size.

\subsubsection{What are the Best Sequence Length, Split Size, and Top-$k$ Values?}
\label{sec:sensitivity_ranker}

    \begin{table*}[p]
        \tiny
        \caption{Avey’s performance under varying sequence lengths (\(N\)), split sizes (\(S\)), and top-\(k\) values. The study trains 69 models with 140M parameters on 10B tokens. The strongest downstream performance is achieved at \(N=512\), \(S=64\), and \(k=7\).}
        \label{tab:sensitivity_ranker}
        \centering
        \resizebox{\textwidth}{!}{%
        \begin{tabular}{cccccccccccc}
        \toprule
        \textbf{$N$} & \textbf{$S$} & \textbf{$k$} &
        \textbf{Perplexity} & \textbf{ARC-C} & \textbf{ARC-E} & \textbf{HellaSwag} &
        \textbf{OBQA} & \textbf{PIQA} & \textbf{SIQA} & \textbf{Winogrande} &
        \textbf{Average} \\
        \midrule
        \multirow{15}{*}{256}
         & \multirow{8}{*}{16}
         & 1  & 43.59 & 21.16 & 37.67 & 30.86 & 27.20 & 64.20 & 37.72 & 51.85 & 38.66 \\
         & & 3  & 31.83 & 22.18 & 39.81 & 32.23 & 26.80 & 63.76 & 38.02 & 51.07 & 39.12 \\
         & & 5  & 27.14 & 24.15 & 39.02 & 32.57 & 26.00 & 64.20 & 38.02 & 53.20 & 39.59 \\
         & & 7  & 25.65 & 22.27 & 38.59 & 31.89 & 28.60 & 64.80 & 37.97 & 50.43 & 39.22 \\
         & & 9  & 24.28 & 23.72 & 38.55 & 32.44 & 28.20 & 64.15 & 37.36 & 51.54 & 39.42 \\
         & & 11 & 22.89 & 22.53 & 39.27 & 32.45 & 27.40 & 65.07 & 38.02 & 51.30 & 39.43 \\
         & & 13 & 23.85 & 22.35 & 38.97 & 31.04 & 26.80 & 64.91 & 37.51 & 49.72 & 38.76 \\
         & & 15 & \textbf{22.04} & 22.61 & 38.80 & 32.66 & 28.60 & 65.61 & 38.54 & 52.25 & \textbf{39.87} \\
        \cmidrule{2-12}
         & \multirow{4}{*}{32}
         & 1  & 36.58 & 23.98 & 39.18 & 32.57 & 26.80 & 64.25 & 37.36 & 50.83 & 39.28 \\
         & & 3  & 31.51 & 22.78 & 39.56 & 33.15 & 25.00 & 64.91 & 38.74 & 52.25 & 39.48 \\
         & & 5  & \textbf{30.06} & 23.38 & 38.76 & 33.51 & 28.60 & 65.34 & 37.87 & 52.41 & \textbf{39.98} \\
         & & 7  & 30.58 & 24.23 & 38.97 & 33.14 & 26.80 & 65.23 & 37.82 & 51.14 & 39.62 \\
        \cmidrule{2-12}
         & \multirow{2}{*}{64}
         & 1  & 32.76 & 23.89 & 39.18 & 33.26 & 27.60 & 66.05 & 38.69 & 50.99 & \textbf{39.95} \\
         & & 3  & \textbf{30.90} & 23.04 & 40.11 & 33.70 & 27.40 & 64.91 & 38.08 & 50.83 & 39.72 \\
        \cmidrule{2-12}
         & 128 & 1 & \textbf{31.30} & 22.27 & 39.65 & 33.07 & 28.20 & 65.45 & 39.20 & 51.78 & \textbf{39.95} \\

         \midrule
         \multirow{23}{*}{512}
          & \multirow{8}{*}{16}
          & 1  & 42.51 & 22.27 & 37.42 & 31.38 & 27.80 & 63.87 & 37.31 & 51.07 & 38.73 \\
          & & 3  & 29.28 & 22.87 & 38.76 & 31.99 & 28.00 & 64.96 & 36.49 & 52.17 & 39.32 \\
          & & 5  & 24.64 & 22.01 & 38.05 & 32.69 & 26.80 & 63.98 & 36.44 & 52.01 & 38.85 \\
          & & 7  & 23.49 & 23.46 & 38.13 & 31.72 & 26.40 & 64.58 & 37.82 & 49.41 & 38.79 \\
          & & 9  & 20.79 & 23.38 & 39.02 & 31.74 & 27.00 & 64.04 & 38.23 & 51.14 & 39.22 \\
          & & 11 & 19.52 & 23.63 & 37.88 & 32.13 & 27.20 & 64.91 & 38.02 & 53.67 & 39.63 \\
          & & 13 & 19.45 & 22.44 & 37.54 & 31.34 & 26.80 & 63.33 & 36.80 & 50.83 & 38.44 \\
          & & 15 & \textbf{17.95} & 21.84 & 36.66 & 31.39 & 27.80 & 64.09 & 37.82 & 51.07 & \textbf{39.87} \\
         \cmidrule{2-12}
          & \multirow{8}{*}{32}
          & 1  & 35.47 & 22.87 & 39.94 & 32.49 & 28.40 & 64.47 & 38.18 & 51.70 & 39.72 \\
          & & 3  & 29.49 & 22.95 & 39.18 & 33.22 & 25.60 & 65.13 & 39.00 & 50.51 & 39.37 \\
          & & 5  & 27.99 & 22.78 & 37.71 & 33.46 & 28.20 & 65.02 & 38.33 & 52.72 & 39.75 \\
          & & 7  & 28.07 & 22.01 & 40.07 & 33.45 & 29.20 & 64.80 & 37.67 & 50.75 & 39.71 \\
          & & 9  & 27.17 & 23.89 & 39.02 & 33.46 & 28.40 & 64.91 & 38.59 & 50.20 & 39.78 \\
          & & 11 & 26.77 & 22.87 & 39.65 & 32.55 & 27.20 & 64.09 & 38.33 & 51.14 & 39.40 \\
          & & 13 & \textbf{25.72} & 23.55 & 38.97 & 33.52 & 29.00 & 65.67 & 37.56 & 51.85 & \textbf{40.02} \\
          & & 15 & 26.29 & 22.70 & 39.23 & 32.53 & 29.40 & 64.80 & 38.08 & 50.04 & 39.54 \\
         \cmidrule{2-12}
          & \multirow{4}{*}{64}
          & 1  & 31.67 & 23.46 & 39.35 & 33.15 & 27.80 & 65.02 & 38.13 & 51.70 & 39.80 \\
          & & 3  & 29.51 & 23.38 & 37.92 & 33.12 & 28.40 & 65.72 & 39.10 & 50.83 & 39.78 \\
          & & 5  & 29.31 & 24.23 & 39.77 & 33.17 & 27.60 & 64.58 & 38.33 & 52.09 & 39.97 \\
          & & 7  & \textbf{28.02} & 24.49 & 39.98 & 33.77 & 29.80 & 65.13 & 38.08 & 51.30 & \textbf{40.36} \\
         \cmidrule{2-12}
          & \multirow{2}{*}{128}
          & 1  & \textbf{29.25} & 23.72 & 39.90 & 33.76 & 28.20 & 64.09 & 37.10 & 50.99 & 39.68 \\
          & & 3  & 29.77 & 22.70 & 39.10 & 33.38 & 28.80 & 65.23 & 38.74 & 51.62 & \textbf{39.94} \\
         \cmidrule{2-12}
          & 256 & 1 & \textbf{29.26} & 22.70 & 39.02 & 33.49 & 27.00 & 64.25 & 37.51 & 52.41 & \textbf{39.48} \\

          \midrule
          \multirow{27}{*}{1024}
           & \multirow{8}{*}{16}
           & 1  & 41.64 & 21.42 & 37.16 & 31.12 & 29.80 & 64.47 & 37.56 & 50.75 & 38.90 \\
           & & 3  & 28.08 & 22.61 & 38.26 & 31.88 & 27.20 & 64.64 & 38.08 & 50.99 & 39.09 \\
           & & 5  & 23.69 & 22.18 & 38.38 & 31.94 & 28.80 & 64.09 & 37.87 & 51.22 & \textbf{39.21} \\
           & & 7  & 21.48 & 23.81 & 38.05 & 31.41 & 27.00 & 63.38 & 36.80 & 50.59 & 38.72 \\
           & & 9  & 19.83 & 22.53 & 37.50 & 31.90 & 26.80 & 64.47 & 37.92 & 49.64 & 38.68 \\
           & & 11 & 18.34 & 21.93 & 37.16 & 31.45 & 28.60 & 65.13 & 37.77 & 50.67 & 38.96 \\
           & & 13 & 16.80 & 23.55 & 37.50 & 30.55 & 26.80 & 63.11 & 36.95 & 52.33 & 38.68 \\
           & & 15 & \textbf{15.33} & 23.29 & 37.54 & 31.04 & 27.60 & 63.06 & 37.77 & 50.91 & 38.74 \\
          \cmidrule{2-12}
           & \multirow{8}{*}{32}
           & 1  & 35.07 & 22.70 & 39.98 & 32.87 & 27.60 & 65.23 & 37.77 & 51.14 & 39.61 \\
           & & 3  & 28.54 & 23.55 & 38.55 & 32.91 & 27.60 & 64.74 & 37.51 & 50.28 & 39.31 \\
           & & 5  & 26.25 & 22.95 & 39.06 & 33.39 & 28.60 & 64.64 & 38.28 & 50.04 & 39.57 \\
           & & 7  & 26.29 & 24.06 & 38.76 & 32.70 & 27.60 & 64.80 & 37.67 & 53.12 & \textbf{39.82} \\
           & & 9  & 24.79 & 23.63 & 38.89 & 33.34 & 27.80 & 64.53 & 37.72 & 52.09 & 39.71 \\
           & & 11 & 24.33 & 21.93 & 38.76 & 32.56 & 26.40 & 64.36 & 37.36 & 51.38 & 38.96 \\
           & & 13 & 23.44 & 22.78 & 37.46 & 32.73 & 29.00 & 65.23 & 37.31 & 50.67 & 39.31 \\
           & & 15 & \textbf{23.14} & 23.72 & 39.56 & 32.39 & 28.40 & 63.60 & 37.31 & 51.38 & 39.48 \\
          \cmidrule{2-12}
           & \multirow{8}{*}{64}
           & 1  & 30.84 & 23.89 & 38.51 & 33.31 & 27.20 & 65.18 & 38.28 & 49.88 & 39.46 \\
           & & 3  & 27.82 & 22.61 & 39.60 & 33.39 & 28.60 & 64.74 & 38.84 & 50.20 & 39.71 \\
           & & 5  & 27.89 & 23.04 & 40.49 & 32.97 & 30.00 & 65.02 & 38.49 & 49.09 & 39.87 \\
           & & 7  & 27.48 & 24.06 & 39.27 & 33.54 & 28.80 & 65.72 & 37.72 & 50.75 & 39.98 \\
           & & 9  & 27.37 & 22.35 & 39.98 & 33.48 & 28.00 & 65.29 & 38.74 & 52.64 & 40.07 \\
           & & 11 & 27.38 & 23.29 & 39.35 & 33.05 & 28.60 & 65.72 & 37.97 & 50.67 & 39.81 \\
           & & 13 & 27.44 & 24.23 & 39.77 & 33.05 & 28.20 & 66.05 & 37.31 & 51.07 & 39.95 \\
           & & 15 & \textbf{26.85} & 24.32 & 40.36 & 34.01 & 28.60 & 65.45 & 37.72 & 51.30 & \textbf{40.25} \\
          \cmidrule{2-12}
           & \multirow{4}{*}{128}
           & 1  & 28.62 & 23.12 & 40.57 & 33.56 & 27.80 & 65.61 & 38.54 & 51.30 & 40.07 \\
           & & 3  & \textbf{27.08} & 23.89 & 39.81 & 33.65 & 29.00 & 64.69 & 37.72 & 52.09 & 40.12 \\
           & & 5  & 28.07 & 23.98 & 40.03 & 33.22 & 29.60 & 65.89 & 39.10 & 50.67 & \textbf{40.35} \\
           & & 7  & 27.27 & 24.32 & 38.85 & 33.92 & 27.40 & 65.13 & 38.13 & 49.88 & 39.66 \\
          \cmidrule{2-12}
           & \multirow{2}{*}{256}
           & 1  & \textbf{28.30} & 22.61 & 38.38 & 33.22 & 27.80 & 64.85 & 38.79 & 51.30 & \textbf{39.56} \\
           & & 3  & 28.53 & 23.04 & 39.90 & 32.68 & 27.40 & 64.36 & 37.62 & 49.88 & 39.27 \\
          \cmidrule{2-12}
           & 512 & 1 & \textbf{28.24} & 24.40 & 39.39 & 32.98 & 28.20 & 65.13 & 37.72 & 50.59 & \textbf{39.77} \\
    
         \bottomrule

        \end{tabular}}
        \end{table*}

We now analyze how Avey’s perplexity and overall performance vary with three key parameters, the ranker’s top-\(k\) selected splits, the split size \(S\), and the sequence length \(N\). We consider \(N \in \{256, 512, 1024\}\). For each \(N\), the split size \(S\) grows geometrically from \(16\), doubling at each step up to the maximum permissible value \(N/2\). Subsequently, for any \((N,S)\), the number of top-ranked splits \(k\) ranges from \(1\) (i.e., contextualizing the current split with one additional relevant split) up to \(N/S - 1\). To keep the experiment count tractable, we vary \(k\) arithmetically from \(1\) to at most \(15\), in increments of \(2\) whenever feasible. All runs use the best configurations established in Sections~\ref{sec:activation_enricher}, \ref{sec:activation_contextualizer}, \ref{sec:best_expansion_factor}, \ref{sec:best_tail_size}, and \ref{sec:embedding_dim}.

As shown in Table~\ref{tab:sensitivity_ranker}, Avey’s perplexity is highest when both \(S\) and \(k\) are very small (e.g., \(S=16\) and \(k=1\)). Although a small \(S\) can help filter out irrelevant embeddings and \emph{denoise} contextualization, pairing it with a very small \(k\) can starve the contextualizer of sufficient signal to form expressive representations\footnote{Recall from Section~\ref{contextualizer} that the context width \(C\) is defined as \(C = S(k+1)\).}. To expand the effective context (i.e., increase \(S(k+1)\)) and enrich the resulting representations, one can increase either \(S\) or \(k\). For example, under \(N=256\) and \(S=16\), increasing \(k\) monotonically reduces perplexity and improves benchmark accuracy. However, a larger context does not invariably yield better outcomes, especially when it admits a higher proportion of irrelevant embeddings. This effect appears when \(k>3\) for \(N=1024\) and \(S=128\), where perplexity rises and downstream accuracy degrades.

Unlike \(S\) and \(k\), the sequence length \(N\) controls the size of the candidate pool from which the ranker selects top-\(k\) splits for the current split during training. A larger \(N\) allows the ranker to reach further back in the sequence history, potentially retrieving more relevant splits and lowering perplexity. For example, increasing \(N\) from \(512\) to \(1024\) while holding \(S=16\) and \(k=15\) fixed reduces perplexity by \(5.4\%\). Nevertheless, lower perplexity does not always translate into higher downstream accuracy. With \(N=512\) and \(S=16\), for instance, \(k=15\) attains the lowest perplexity, yet \(k=5\) yields better benchmark accuracy. As noted in Section~\ref{sec:activation_enricher}, loss remains a proxy for overall modeling capability and does not always precisely predict downstream performance. As shown in Table~\ref{tab:sensitivity_ranker}, the strongest empirical results are obtained with \(N=512\), \(S=64\), and \(k=7\), which we accordingly adopt as Avey’s default configuration.

\subsubsection{RMSNorm or LayerNorm}
\label{sec:norm_study}

\begin{table*}[ht]
\caption{Avey’s performance with RMSNorm versus LayerNorm. The study trains a 153M-parameter model on 10B tokens using the full architecture (neural processor + ranker) with the best configuration from Table~\ref{tab:sensitivity_ranker}.}
\label{tab:normalization_techniques}
\centering
\resizebox{\textwidth}{!}{%
\begin{tabular}{lccccccccc}
\toprule
\textbf{Normalization Method} & \textbf{Perplexity} & \textbf{ARC-C} & \textbf{ARC-E} & \textbf{HellaSwag} &
\textbf{OBQA} & \textbf{PIQA} & \textbf{SIQA} & \textbf{Winogrande} & \textbf{Average} \\
\midrule
RMSNorm   & \textbf{28.02} & 24.49 & 39.98 & 33.77 & 29.8 & 65.13 & 38.08 & 51.30 & \textbf{40.36} \\
LayerNorm & 30.93          & 23.55 & 39.65 & 33.24 & 28.8 & 65.07 & 38.28 & 49.57 & 39.74 \\
\bottomrule
\end{tabular}}
\end{table*}

In all preceding studies, we employed RMSNorm~\citep{zhang2019rmsnorm} as the normalization method. We now compare Avey against an alternative standard, that is, LayerNorm~\citep{ba2016layer}. Recall that Avey applies pre-layer normalization to the input embeddings at every layer (as illustrated in Fig.~\ref{fig:neural_processor}) and a {\em single} post-layer normalization to the output embeddings after the final layer, immediately before token prediction.

We evaluate each normalization method using Avey’s full architecture, comprising both the neural processor and the ranker. The ranker is fixed to the best-performing setting from Section~\ref{sec:sensitivity_ranker} (i.e., sequence length \(N=512\), split size \(S=64\), and top-\(k=7\)). The neural processor adopts the optimal configurations identified in Sections~\ref{sec:activation_enricher}, \ref{sec:activation_contextualizer}, \ref{sec:best_expansion_factor}, \ref{sec:best_tail_size}, and \ref{sec:embedding_dim}. As before, we train a 153M-parameter model on 10B tokens from the FineWeb dataset. Results are summarized in Table~\ref{tab:normalization_techniques}. On average, RMSNorm slightly outperforms LayerNorm and is subsequently used in Avey by default.

\subsubsection{Learning Rate: To Decay or Not to Decay?}
\label{sec:lr_schedules}

\begin{table*}[ht]
\centering
\small
\caption{Avey’s performance under two learning-rate schedules, constant and cosine decay, across varying peak learning rates. All models use the full architecture (neural processor + ranker) with the best configuration from Table~\ref{tab:sensitivity_ranker}, and are trained with 153M parameters on 10B tokens.}
\label{tab:lr_schedule_comparison}
\setlength{\tabcolsep}{4pt}
\resizebox{\textwidth}{!}{%
\begin{tabular}{llccccccccc}
\toprule
\textbf{Schedule} & \textbf{LR} & \textbf{Perplexity} & \textbf{ARC-C} & \textbf{ARC-E} & \textbf{HellaSwag} &
\textbf{OBQA} & \textbf{PIQA} & \textbf{SIQA} & \textbf{Winogrande} & \textbf{Average} \\
\midrule
\multirow{3}{*}{Constant}
& 8e-04 & 28.21 & 23.12 & 39.81 & 33.52 & 28.4 & 64.96 & 38.13 & 52.41 & 40.05 \\
& 1e-03 & 27.35 & 24.06 & 40.32 & 33.88 & 29.6 & 65.13 & 38.54 & 51.46 & \textbf{40.43} \\
& 3e-03 & 30.38 & 23.81 & 38.34 & 32.90 & 28.2 & 63.38 & 37.77 & 52.33 & 39.53 \\
\midrule
\multirow{3}{*}{Cosine Decay}
& 6e-04 & 26.24 & 22.87 & 40.95 & 33.76 & 29.2 & 65.02 & 37.72 & 49.01 & 39.79 \\
& 8e-04 & 25.64 & 24.06 & 39.31 & 34.43 & 29.6 & 65.83 & 37.87 & 49.80 & 40.13 \\
& 1e-03 & \textbf{25.00} & 23.21 & 41.12 & 34.76 & 27.0 & 65.67 & 38.38 & 50.75 & \textbf{40.13} \\
\bottomrule
\end{tabular}}
\end{table*}

In all prior experiments, we used a constant learning rate of $1e{-3}$. In this study, we assess Avey’s performance under varying peak learning rates and learning-rate schedules. Specifically, we compare a cosine-decay schedule against a constant schedule (i.e., cosine decay with effectively infinite cycle length, yielding no decay). The study adopts the experimental configurations established in Sections~\ref{sec:activation_enricher}, \ref{sec:activation_contextualizer}, \ref{sec:best_expansion_factor}, \ref{sec:best_tail_size}, and \ref{sec:embedding_dim}. In addition, we utilize Avey’s full architecture, encompassing the neural processor and ranker, with the best configuration from Table~\ref{tab:sensitivity_ranker}.

As shown in Table~\ref{tab:lr_schedule_comparison}, cosine decay consistently achieves lower loss than a constant schedule across the learning rates tested. This trend aligns with the \emph{Chinchilla} findings~\citep{hoffmann2022training}, which report that when the cosine cycle length substantially exceeds the total number of training steps (by at least \(25\%\)), performance degrades. Notably, a constant schedule corresponds to an effectively infinite cycle length. In contrast, setting the cycle length to approximately match the training duration yields the best final loss~\citep{hoffmann2022training}.

To this end, we adopt a cosine-decay schedule with a peak learning rate of \(1e{-3}\)  for Avey, as it yields the lowest training loss across all the runs. We note, however, that Table~\ref{tab:lr_schedule_comparison} also shows a mismatch between loss and downstream accuracy. For example, a higher loss of \(3.308\) under the constant schedule produced slightly better benchmark performance than a lower loss of \(3.218\) under cosine decay at the same peak rate. While constant learning rates can be effective for short or exploratory runs (this study uses 10B tokens), it is generally the case that, as the number of training tokens increases, the learning rate should decrease to allow the optimizer to \emph{settle} into a lower-loss region~\citep{you2019does}. Consequently, schedules with decay are typically preferred for longer or large-scale training~\citep{hoffmann2022training,bergsma2025straight}.

\subsection{Ablation Study}
\label{sec:ablation}

\begin{table*}[ht]
\centering
\tiny
\caption{Ablation results comparing Avey variants, with individual components removed or substituted.}
\label{tab:ablation_avey}
\setlength{\tabcolsep}{4pt}
\begin{tabularx}{\textwidth}{l *{9}{>{\centering\arraybackslash}X}}
\toprule
\textbf{Model Variant} & \textbf{Perplexity} & \textbf{ARC-C} & \textbf{ARC-E} & \textbf{Hella} & \textbf{OBQA} & \textbf{PIQA} & \textbf{SIQA} & \textbf{Wino} & \textbf{Average} \\
\midrule
Avey full (all features)                                 & 30.00 & 25.17 & 39.90 & 33.59 & 28.8  & 65.56 & 37.62 & 51.62 & \textbf{40.32} \\
Avey \emph{without} dynamic parameterization             & 34.31 & 25.00 & 40.66 & 32.99 & 28.8  & 65.34 & 36.64 & 50.51 & 39.99 \\
Avey \emph{without} bypassing                            & 32.55 & 22.61 & 38.38 & 32.31 & 28.0  & 64.20 & 38.28 & 52.09 & 39.41 \\
Avey \emph{without} embedding expansion                  & 39.94 & 22.44 & 37.92 & 28.75 & 25.4  & 62.40 & 38.64 & 52.01 & 38.22 \\
Avey \emph{without} weighting selected splits            & 31.17 & 22.78 & 38.55 & 33.25 & 28.0  & 65.89 & 37.82 & 52.09 & 39.77 \\
Avey \emph{without} the ranker                           & 29.48 & 23.72 & 38.59 & 32.52 & 28.0  & 63.66 & 37.67 & 53.20 & 39.62 \\
Avey \emph{with} self-attention in place of neural proc. & 31.39 & 22.61 & 39.27 & 31.99 & 28.0  & 64.58 & 38.33 & 51.38 & 39.45 \\
\bottomrule
\end{tabularx}
\end{table*}

In this section, we conduct a series of ablations on Avey’s core components, building on the best configurations identified in Section~\ref{sec:design_choices}. All experiments use Avey’s full architecture, comprising the neural processor and ranker, with the \emph{small} model (153M parameters) as the baseline. We train this model on 10B tokens from the FineWeb dataset, following the training methodology and hyperparameters detailed in Section~\ref{sec:experimental_methodology}.

We begin by examining the effect of \emph{dynamic parameterization} on both perplexity and downstream benchmark performance. As described in Section~\ref{contextualizer}, dynamic parameterization enables each neuron in the contextualizer to compute a cosine similarity between its input embedding and the embeddings associated with other neurons, weight those embeddings accordingly, and aggregate them via a learned weighted sum. This mechanism induces \emph{selectivity}~\citep{gu2023mamba} in the neural processor, making its parameterization input-dependent. Table~\ref{tab:ablation_avey} summarizes the results. Disabling this component increases perplexity by \(14.3\%\) and reduces average downstream performance by \(0.8\%\), highlighting its importance.

Second, we assess the impact of \emph{partial-embedding bypassing} introduced in Section~\ref{enricher}. This mechanism forwards a fraction of each expanded embedding directly to the fuser, preserving raw, distinctive features that can promote more diverse representations. As reported in Table~\ref{tab:ablation_avey}, removing this component increases perplexity by \(8.5\%\) and reduces average downstream performance by \(2.2\%\), underscoring its significance.

Third, we set the enricher’s expansion factor to \(1\), effectively disabling embedding expansion. As illustrated in Table~\ref{tab:ablation_avey}, this change increases perplexity by \(33.1\%\) and reduces average downstream performance by \(5.2\%\), corroborating the critical role of embedding expansion in the model’s effectiveness.

Fourth, we remove the weighting of each selected split by its corresponding normalized MaxSim score, thereby preventing the contextualizer from scaling each split’s contribution during contextualization. As depicted in Table~\ref{tab:ablation_avey}, this adjustment increases perplexity by \(3.8\%\) and reduces average downstream performance by \(1.37\%\), indicating the importance of this technique.

Fifth, we evaluate Avey \emph{without} the ranker to assess its impact on downstream performance beyond its primary role of enabling effective extrapolation outside the trained context window. As shown in Table~\ref{tab:ablation_avey}, the ranker improves the neural processor’s performance, primarily by enhancing contextualization through more meaningful cross-token interactions. We note, however, a slight increase in loss (\(0.5\%\)), which again highlights (as in Section~\ref{sec:activation_enricher}) the discrepancy between the objectives of pertaining and downstream tasks.

Finally, we replace Avey’s neural processor with a self-attention module to assess the relative contribution of each component to overall performance, given that both are designed to facilitate cross-token interactions. As illustrated in Table~\ref{tab:ablation_avey}, this substitution increases perplexity by \(4.6\%\) and reduces average downstream performance by \(2.1\%\), underscoring the centrality of the neural processor and suggesting that self-attention is less effective within Avey’s architectural framework.

\subsection{Short-Range Benchmark Results}
\label{sec:short_range}

\begin{table*}[ht]
\centering
\small
\caption{Zero-shot performance across multiple NLP tasks.}
\label{tab:zero_shot_results}
\setlength{\tabcolsep}{4pt}  
\resizebox{\textwidth}{!}{%
\begin{tabular}{lcccccccc}
\toprule
\textbf{Model} & \textbf{ARC-C} & \textbf{ARC-E} & \textbf{HellaSwag} & \textbf{OBQA} & \textbf{PIQA} & \textbf{SIQA} & \textbf{Winogrande} & \textbf{Average} \\
\midrule
\textbf{Avey-153M}             & \textbf{24.37} & 42.33 & 39.36 & \textbf{31.40} & 68.37 & 39.13 & 51.28 & 42.32 \\
Transformer++-152M            & 23.63 & 43.17 & 39.32 & 29.80 & 67.01 & 38.89 & 50.22 & 41.72 \\
Mamba-144M                    & 24.17 & \textbf{43.53} & 40.55 & 30.40 & 68.32 & \textbf{39.41} & \textbf{52.72} & \textbf{42.73} \\
RWKV-7-168M                   & 24.17 & 43.01 & \textbf{41.55} & 29.67 & \textbf{68.72} & 39.17 & 51.09 & 42.48 \\
\midrule
\textbf{Avey-496M}            & 27.50 & 48.95 & 51.82 & 32.47 & 72.49 & 40.15 & 54.38 & 46.82 \\
Transformer++-488M           & 26.73 & 48.09 & 52.66 & 31.73 & 72.13 & 39.93 & 55.25 & 46.65 \\
Mamba-500M                   & \textbf{28.64} & \textbf{51.02} & 54.15 & 34.47 & 73.03 & \textbf{40.84} & 55.49 & \textbf{48.23} \\
RWKV-7-501M                  & 27.13 & 49.37 & \textbf{54.54} & \textbf{36.27} & \textbf{73.58} & 39.40 & \textbf{55.72} & 48.00 \\
\midrule
\textbf{Avey-1.52B}           & 31.26 & 56.55 & 61.42 & 36.80 & 75.61 & 42.00 & 57.06 & 51.53 \\
Transformer++-1.5B           & 30.00 & 56.29 & 63.87 & \textbf{38.00} & 76.01 & \textbf{42.24} & \textbf{61.38} & 52.54 \\
Mamba-1.4B                   & 32.13 & 57.74 & 63.74 & 36.85 & 76.19 & 42.00 & 60.40 & 52.72 \\
RWKV-7-1.5B                  & \textbf{32.94} & \textbf{59.05} & \textbf{64.43} & 37.13 & \textbf{76.84} & 41.71 & 60.06 & \textbf{53.17} \\
\bottomrule
\end{tabular}}
\vspace{-0.8em}
\end{table*}

In this section, we evaluate Avey on standard autoregressive language-modeling benchmarks and compare it against Transformer++, Mamba, and RWKV\text{-}7 across three parameter scales, \emph{small}, \emph{medium}, and \emph{large} (see Table~\ref{models_params}). We adopt a suite of widely used zero-shot downstream tasks, as outlined in Section~\ref{sec:experimental_methodology}. Results are summarized in Table~\ref{tab:zero_shot_results}. For the \emph{small} models, Avey, Mamba, and RWKV\text{-}7 outperform Transformer++ by average margins of \(1.43\%\), \(2.41\%\), and \(1.82\%\), respectively. Additionally, Mamba and RWKV\text{-}7 slightly exceed Avey by \(0.90\%\) and \(0.30\%\). For the \emph{medium} models, Avey, Mamba, and RWKV\text{-}7 again surpass Transformer++ by \(0.30\%\), \(3.40\%\), and \(2.90\%\), respectively. For the \emph{large} models, Avey underperforms Transformer++ by \(1.90\%\), while Mamba and RWKV\text{-}7 edge out Avey by \(0.71\%\) and \(1.19\%\), respectively.

The results above assume a fixed training budget of 100B tokens. To analyze Avey’s scaling behavior with respect to model size, we conduct additional experiments following the \emph{Chinchilla} scaling laws~\citep{hoffmann2022training}, which recommend increasing the number of training tokens in proportion to model size. Accordingly, we adjust the training steps and token budgets to align with these prescriptions. All results are reported in Appendix~\ref{sec:scaling_laws}, which also details the model configurations (e.g., number of layers and embedding dimensions), optimization settings (e.g., learning rates), training schedules, and total tokens for all evaluated models. Our methodology largely follows~\citet{gu2023mamba}, with minor modifications to satisfy parameter-budget constraints. Results in Appendix~\ref{sec:scaling_laws} show that Avey exhibits the strongest scaling behavior, followed by Transformer++, Mamba, and RWKV-7, in that order.

\subsection{Long-Range Benchmark Results}
\label{sec:long_range}

\begin{figure}[ht]
  \centering
  \includegraphics[width=\linewidth, height=4cm, keepaspectratio=true]{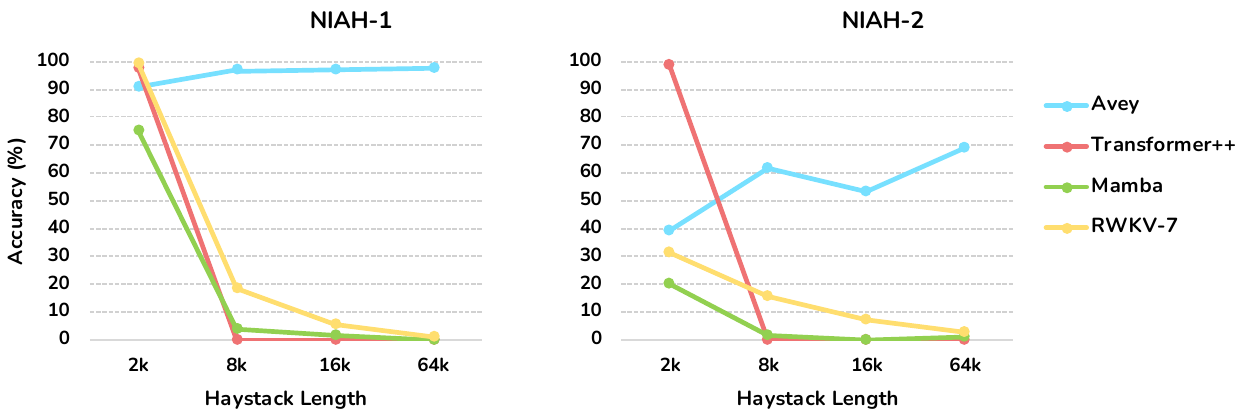}
  \caption{Performance comparison between Transformer++, Mamba, RWKV-7, and Avey on S-NIAH-1 and S-NIAH-2. The x-axis denotes the lengths of haystacks (i.e., documents with distractor texts, varying from 2k to 64k tokens). All models use 0.5B parameters. Similar results are shown in Appendix~\ref{sec:extra_long_range_results} for other model sizes.}
  \label{fig:NiaH_1_and_2}
  \vspace{-1.1em}
\end{figure}

We now evaluate Avey, Transformer++, Mamba, and RWKV\text{-}7 on benchmarks targeting long-range dependencies. Specifically, we use the standard Single Needle\text{-}In\text{-}A\text{-}Haystack (S\text{-}NIAH) suite from RULER~\citep{hsieh2024ruler}, as described in Section~\ref{sec:experimental_methodology}. The suite includes multiple variants, notably S\text{-}NIAH\text{-}1 and S\text{-}NIAH\text{-}2. S\text{-}NIAH\text{-}1 requires retrieving the value associated with a given key (the \emph{pass\text{-}key}) from a distractor context (the \emph{haystack}) containing many key–value pairs. The pass\text{-}key (or the \emph{needle}) appears exactly once, and the model must recover its corresponding value regardless of position within the haystack. S\text{-}NIAH\text{-}2 follows the same structure but is more challenging, whereby the target value is numeric (e.g., a random 9\text{-}digit number), demanding \emph{exact} recall in which any single\text{-}digit error is scored as incorrect. Consequently, this variant tests precise extraction of structured numerical information from long contexts.

Fig.~\ref{fig:NiaH_1_and_2} demonstrates the results for Avey, Transformer++, Mamba, and RWKV\text{-}7 on both S\text{-}NIAH\text{-}1 and S\text{-}NIAH\text{-}2. As noted in Section~\ref{sec:experimental_methodology}, Transformer++, Mamba, and RWKV\text{-}7 were all trained with a 2{,}048\text{-}token context window. Transformer++ performs strongly on both benchmarks while the haystack length remains within this window. Once the haystack length exceeds the window width, the accuracy of Transformer++ collapses. Mamba and RWKV\text{-}7 generalize somewhat beyond their training windows, but their performance also degrades substantially as the haystack grows. In contrast, Avey maintains high accuracy on both benchmarks \emph{despite being trained with a 512\text{-}token context window}. For example, on S\text{-}NIAH\text{-}2 with a 64k\text{-}token haystack, Avey outperforms Mamba and RWKV\text{-}7 by average margins of \(85.25\%\) and \(23.6\%\), respectively. On S\text{-}NIAH\text{-}1 under the same 64k\text{-}token setting, Avey attains \(97.8\%\) accuracy, whereas Mamba and RWKV\text{-}7 drop to \(0\%\) and \(0.8\%\), respectively.

Interestingly, Avey’s performance often improves as the haystack length increases, highlighting strong extrapolative capability. One plausible explanation is that as the haystack (or the sequence length \(N\)) grows, the candidate pool from which the ranker selects top-\(k\) splits also expands. As discussed in Section~\ref{sec:sensitivity_ranker}, a larger \(N\) enables the ranker to retrieve more relevant splits while discarding less informative ones, improving the quality of contextualization and potentially boosting accuracy. This interpretation is consistent with findings in Section~\ref{sec:ablation}, where introducing the ranker yielded measurable gains. Notably, embeddings containing a needle (whether in S\text{-}NIAH\text{-}1 or S\text{-}NIAH\text{-}2) are not processed in isolation but are contextualized alongside other embeddings. As such, better ranker-driven context selection can result in more precise value recall. That said, whether this fully explains Avey’s increasing accuracy with longer haystacks remains an open question, and further interpretability studies are needed to better understand the underlying drivers of this behavior.

\section{Related Work}
\label{sec:r_w}


Recurrent Neural Networks (RNNs)~\citep{elman1990finding, rumelhart1986learning} are designed to process sequential data by capturing temporal dependencies, making them well-suited for tasks where input order is essential. However, their cyclical nature limits their potential for parallel computation and exposes them to vanishing and exploding gradient problems. As a result, they typically struggle to effectively learn long-range dependencies. While architectures like Long Short-Term Memory (LSTM)~\citep{hochreiter1997long} and Gated Recurrent Units (GRU)~\citep{cho2014learning} mitigate these gradient-related issues, they remain slow to optimize and challenging to scale due to retaining RNN’s core recurrent and sequential structure.

In contrast, the Transformer~\citep{vaswani2017attention} employs a self-attention mechanism to process each sequence of tokens simultaneously. More precisely, it promotes two key design principles: (1) a recurrent-free architecture, which enables parallel computation of token embeddings, while still capturing their order through positional encodings, and (2) a multi-head self-attention approach, which facilitates cross-token interactions to further enrich the expressiveness of embeddings. These innovations make the Transformer highly effective, as well as parallelizable and efficient to train. However, they also cause its computational and memory requirements to scale quadratically with sequence length, making it expensive and less efficient for very long sequences.

To address the Transformer's quadratic computation and memory costs, a wide array of approaches have been proposed, including linear attention~\citep{kitaev2020reformer, katharopoulos2020transformers, choromanski2020rethinking, peng2021random, zhai2021attention}, sparse or local attention~\citep{yuan2025native, child2019generating, parmar2018image}, context compression~\citep{rae2019compressive, wang2020linformer, sukhbaatar2019adaptive, roy2021efficient}, and modified attention computations~\citep{tay2021synthesizer, wu2019pay, tay2020sparse}, to mention just a few. Notably, the Attention-Free Transformer (AFT)~\citep{zhai2021attention} offers a linear drop-in replacement for the quadratic self-attention mechanism. In particular, it weights key and value vectors using learned positional biases and integrates them with query vectors via element-wise multiplication. As such, it eliminates the need to compute and store the costly attention matrix while still preserving global query-value interactions—without requiring architectural modifications or additional tuning. Furthermore, AFT introduces variants such as AFT-local and AFT-conv, which leverage local attention patterns to reduce parameter count and further improve computational and memory efficiency.

RWKV-4~\citep{peng2023rwkv} (the first 3 versions were experimental~\citep{li2024survey}) capitalizes on AFT and suggests combining the strengths of both Transformers and RNNs. To elaborate, unlike Transformers and akin to RNNs, it does not process each input token solely based on its own embedding, but rather as a weighted sum of its embedding and that of the preceding one. To the contrary of traditional RNNs and similar to Transformers, it adopts self-attention, but an extended version of it, namely, that of AFT. This hybrid approach allows RWKV-4 to maintain some of the modeling capabilities of RNNs, while benefiting from the parallelization and scalability features of Transformers.

More precisely, RWKV-4 extends AFT in two distinct ways: (1) it introduces an additional parameter to handle each current token independently, and (2) it implements a per-time-step decay mechanism that selectively removes older content from the recurrent hidden state in a data-dependent manner. This decay mechanism addresses a central limitation of linear attention, which pertains to the  lack of a systematic way to discard outdated information~\citep{schlag2021linear, yang2023gated}.

Architecturally, RWKV-4 consists of homogeneous stacked residual blocks, each encompassing two units, a time-mixing and a channel-mixing ones. The time-mixing unit applies linear attention across tokens, while the channel-mixing unit integrates each element of the current token’s embedding with its corresponding element from the preceding token embedding, leveraging the output of the time-mixing unit.

RWKV-5~\citep{peng2024eagle} enhances RWKV-4’s architecture and learning decay mechanism by replacing traditional vector-valued states with more expressive multi-head, matrix-valued ones. Moreover, it reconfigures receptive states, incorporates supplementary gating mechanisms, and dynamically learns the linear interpolation between the current and preceding token embeddings instead of relying on pre-defined hyperparameters. 

RWKV-6~\citep{peng2024eagle} promotes a new application of low-rank adaptation functions~\citep{hu2022lora, li2024survey}. Specifically, it makes the linear interpolation between the current and preceding tokens data-dependent to improve the selectivity of the model in retaining and discarding information. Additionally, it replaces the static decay mechanism with a dynamic one, allowing each element in the decay vector to fluctuate independently over time in response to the input.

The decay strategies of RWKV-4, RWKV-5, and RWKV-6 still cannot remove values stored at specific keys. DeltaNet~\citep{schlag2021linear} overcomes this drawback by partially replacing the values stored at current keys with equivalent new values, enabling models to erase outdated memories and include up-to-date ones on a per-key basis. However, it only allows a fixed scalar fraction of a value to be replaced from a state via an in-context learning rate parameter, thus demonstrating rigidity in adapting to varying data contexts. 

RWKV-7~\citep{peng2025rwkv} builds upon the principles of DeltaNet and introduces a vector-valued in-context learning rate instead of a scalar-valued one. This allows selective replacement of state data on a channel-wise basis. Furthermore, RWKV-7 employs a vector-valued decay mechanism and uses additional low-rank projections to optimize the trade-off between the number of parameters, computational efficiency, and downstream performance. Lastly, it incorporates Value Residual Learning~\citep{zhou2024value}, which improves the propagation of initial local information via utilizing a residual connection between the value vectors of the current layer and those of the {\em first} layer prior to the attention operation, resulting in enhanced language modeling performance.

Most recently, RWKV-X~\citep{hou2025rwkv} proposed combining the strengths of RWKV and sparse attention, drawing inspiration from Mixture of Block Attention (MoBA)~\citep{lu2025moba}. In particular, RWKV-X restricts each query to attend only to a small, relevant subset of the input, thus reducing computational cost and facilitating the modeling of longer-range dependencies. More precisely, rather than allowing each token to attend to every other token in the sequence (as in traditional self-attention), it constrains each token’s attention to a limited subset (hence, making it sparse), while maintaining the coupling between the input sequence and context window.

Similar to RWKV, RetNet~\citep{sun2023retentive} adopts linear attention and promotes a hybrid approach that blends Transformer- and RNN-like representations, yet with a decay-based memory unit. Specifically, it divides the input sequence into chunks, wherein the Transformer-like parallel representation is applied. Additionally, it enables propagating information sequentially across chunks using the RNN-like representation. Lastly, it uses multiple attention heads, each governed by a distinct decay rate, and replaces LayerNorm~\citep{ba2016layer} with GroupNorm~\citep{wu2018group}.

Although linear attention has been proposed as a promising alternative to quadratic softmax attention~\citep{katharopoulos2020transformers, choromanski2020rethinking, kasai2021finetuning, peng2021random}, existing implementations of it are in practice slower than optimized versions of softmax attention~\citep{dao2022flashattention, dao2023flashattention, yang2023gated}. From an accuracy standpoint, linear attention generally underperforms conventional softmax attention, sometimes by a significant margin in language modeling~\citep{kasai2021finetuning, yang2023gated}.

To this end, and in light of the exponentially growing complexity associated with overcoming the limitations of Transformer-based architectures, there has been a renewed interest in RNN-based alternatives in recent years. Notably, Structured State Space Sequence (S4) models~\citep{gu2021efficiently, gu2021combining}, inspired by the classical state space models (SSMs)~\citep{kalman1960new}, have emerged as a promising paradigm for sequence modeling. These models describe the temporal evolution of a system using differential equations, offering a continuous-time formulation of dynamics, and can be viewed as generalized versions of RNNs.

An SSM as a concept has a broad meaning, which simply refers to the notion of any recurrent process with a latent state~\citep{gu2023mamba}. From this perspective, the RNN-like linear attention model proposed and formulated by~\citep{katharopoulos2020transformers} can be interpreted as a degenerate linear SSM. Interestingly, this justifies the usage of a decay factor in RetNet and RWKV, especially that a decay term (or a forget gate) has been shown to be crucial in RNNs~\citep{hochreiter1997long, van2018unreasonable, cho2014learning}.

Numerous variants of SSMs~\citep{gu2021efficiently, gu2022parameterization, gupta2022diagonal, li2024makes, ma2022mega, orvieto2023resurrecting, smith2022simplified} have demonstrated strong performance across a range of domains, including audio and vision~\citep{goel2022s, nguyen2022s4nd, saon2023diagonal}. Nonetheless, these variants have struggled with language modeling, often lagging behind Transformers by several points in perplexity~\citep{gu2021efficiently}.

From an efficiency standpoint, however, SSMs have shown encouraging results in language modeling. For instance, S4~\citep{gu2021efficiently, gu2021combining}, a prominent SSM, converts the continuous-time state update equation of SSMs into a discrete form, hence, enabling parallel sequence modeling. Moreover, it utilizes the HiPPO (High-Order Polynomial Projection Operator) initialization~\citep{gu2020hippo}, which alleviates the vanishing gradient problem and facilitates processing longer sequences. 

Another example of SSMs is H3~\citep{fu2022hungry}, which improves language modeling by allowing both, the recall of earlier tokens and token-wise comparisons within a sequence. H3 extends S4 by suggesting a state-passing algorithm that enhances computational efficiency on modern hardware. This advancement reduces the hardware-related barriers that have traditionally limited the scalability of SSM-based architectures.

Hyena~\citep{poli2023hyena} capitalizes on H3 by replacing its S4 layer with an MLP-parameterized global convolution~\citep{romero2102ckconv}. S5~\citep{smith2022simplified} proposes using parallel scan~\citep{martin2017parallelizing} to parallelize S4. Liquid S4~\citep{hasani2022liquid} augments S4 with an input-dependent state transition matrix, computed convolutionally in the frequency domain (which is computationally efficient) and mapped back to the time domain using an inverse Fourier transformation. SGConv~\citep{li2024makes}, LongConv~\citep{fu2023simple}, MultiresConv~\citep{shi2023sequence}, and Toeplitz Neural Network~\citep{qin2023toeplitz} all focus on the convolutional representation of S4 as well, aiming to enhance its efficiency~\citep{gu2023mamba}.

Most recently, Mamba~\citep{gu2023mamba} introduced a new class of SSMs known as {\em selective} SSMs, specifically designed to improve the performance of language modeling tasks. Mamba addresses a key limitation in SSMs, namely, their inability to selectively process inputs in an input-dependent manner (i.e., focus on or ignore specific parts of the input sequence). Consequently, it makes the SSM parameters input-dependent, but introduces a technical challenge since traditional SSMs are inherently designed to be time- and input-invariant to ensure computational efficiency. To overcome this challenge, Mamba proposes a hardware-efficient parallel scan (or prefix sum) algorithm~\citep{blelloch1990prefix}, which enables recurrent-style computation without explicitly materializing the expanded state. This design precludes costly I/O operations across GPU memory hierarchies and accelerates both, training and inference. 

Tri Dao and Albert Gu~\citep{dao2024transformers} argue that various approaches to operating SSMs can be reframed as matrix multiplication algorithms involving a specific class of structured matrices known as semiseparable matrices. They further leverage the language of tensor contractions to prove the recurrent formulation of linear attention as proposed by~\citep{katharopoulos2020transformers}, before generalizing it to a new family of structured masked attention (SMA). 

Subsequently, Tri Dao and Albert Gu demonstrated that SSMs and Transformers are fundamentally connected, governed by the mathematical framework of semiseparable matrices and SMA. Additionally, they developed a rich state space duality (SSD) framework of theoretical connections between SSMs and various forms of attention. This framework facilitated the design of Mamba-2, an extended version of Mamba, which aims to improve its efficiency (not performance). Mamba-2 utilizes a scalar data-dependent gating mechanism (like the ones proposed by~\citep{peng2021random, sun2023retentive, beck2024xlstm}), which enables transforming its recurrent structure into a matrix-multiply form, thus allowing for efficient execution on tensor cores and better support for larger hidden state sizes.

The strategy of the SSD framework mirrors that of linear attention~\citep{katharopoulos2020transformers}, which established a connection between autoregressive attention mechanisms and linear RNNs via showing an equivalence between "dual forms" of quadratic kernelized attention and a specific type of linear recurrence. Conceptually, the SSD framework seeks to transfer algorithmic and systems-level optimizations originally developed for Transformers to the realm of SSMs. Its overarching goal is to enable the development of architectures that outperform Transformers, while scaling more efficiently with sequence length.

Finally, several works, including Tolstikhin {\em et al.}~\citep{tolstikhin2021mlp}, Melas-Kyriazi~\citep{melas2021you}, Touvron {\em et al.}~\citep{touvron2022resmlp}, and Ding {\em et al.}~\citep{ding2021repmlp}, among others, have questioned the necessity of self-attention, particularly in the context of Vision Transformers. In contrast, Liu {\em et al.}~\citep{liu2021pay} introduced gMLP, an MLP-based alternative to BERT-style Transformers~\citep{devlin2019bert} that (partially) eliminates self-attention but ultimately underperforms average performance on downstream NLP tasks. gMLP encompasses channel (hidden) and spatial (cross-token) projections with multiplicative gating and {\em static} parameterization. Its gating mechanism is reminiscent of Gated Linear Units (GLUs)~\citep{dauphin2017language, shazeer2020glu, wu2019pay}, as well as earlier architectures such as Highway Networks~\citep{srivastava2015highway} and LSTM-RNNs~\citep{hochreiter1997long}. A key distinction, however, is that gMLP applies gating on the spatially projected dimension and not the hidden one. The gated embedding-wise neural network in Avey’s contextualizer draws inspiration from gMLP.

Unlike all previously mentioned models, Avey abandons self-attention and recurrence, introducing a new architecture composed of a ranker and a dynamically parameterized neural processor. The ranker identifies the most relevant tokens for contextualization, while the neural processor contextualizes them data-dependently. This design decouples sequence length from context width, enabling efficient processing of arbitrarily long sequences without diminishing the influence of distant yet important tokens.

At the core of Avey’s architecture is a weighted-selective-split interaction mechanism, which filters out irrelevant tokens beyond the context window and enables direct interactions only with relevant ones, thus preserving their influence irrespective of sequence length. In addition, Avey employs a partial-embedding bypassing technique that retains a portion of each token’s raw, distinctive features before fusing them with its contextualized ones through a neural network. This technique boosts the performance of Avey (as shown in Appendix~\ref{sec:ablation}) and might help mitigate issues such as entropy collapse~\citep{zhai2023stabilizing} and over-smoothing~\citep{zhou2021deepvit, shi2022revisiting}, especially at large-scale, when the depth of the model is increased significantly.

\section{Conclusion}
\label{sec:conclusion}

In this paper, we introduce Avey, a new foundational architecture for autoregressive language modeling. Unlike traditional models, Avey relies on neither recurrence nor attention. Instead, it employs a neural mechanism to enrich and contextualize embeddings, coupled with a ranker that enables flexible and effective handling of arbitrarily long sequences despite being trained with a small context window. We hope this work lays the groundwork for future research and inspires a promising new direction for scalable, effective language modeling.

\section*{Acknowledgments}
\label{sec:acknowledgments}

We gratefully acknowledge the Ministry of Communications and Information Technology (MCIT) in Qatar and NVIDIA for providing the computational resources that made many of the experiments reported herein possible. We also thank Microsoft Qatar for technical support on the GPU infrastructure provided by MCIT.



\bibliographystyle{plainnat}  
\bibliography{iclr2026_conference}





\appendix

\section{Scaling Laws}
\label{sec:scaling_laws}

\begin{table*}[ht]
\centering
\caption{Model configurations used in the scaling law experiments. Each model is trained at three different sizes and numbers of training tokens increased proportionally, following the Chinchilla scaling laws.}
\label{tab:scaling_config}
\setlength{\tabcolsep}{4pt}
\begin{tabularx}{\textwidth}{lcccc}
\toprule
\textbf{Model} & \textbf{\# Layers (\# Heads)} & \textbf{Embedding Dim.} & \textbf{Learning Rate} & \textbf{\# Tokens} \\
\midrule
Avey-153M              & 26         & 768  & 1.00e-03 & 2B  \\
Avey-496M              & 104        & 768  & 1.00e-03 & 7B  \\
Avey-1.5B              & 48         & 2048 & 1.00e-03 & 20B \\
Transformer++-152M     & 12 (12)    & 768  & 3.00e-03 & 2B  \\
Transformer++-488M     & 26 (16)    & 1024 & 1.50e-03 & 7B  \\
Transformer++-1.5B     & 32 (16)    & 1664 & 1.25e-03 & 20B \\
Mamba-153M             & 28         & 768  & 3.00e-03 & 2B  \\
Mamba-496M             & 42         & 1280 & 1.50e-03 & 7B  \\
Mamba-1.5B             & 52         & 2048 & 1.00e-04 & 20B \\
RWKV-7-152M            & 12         & 768  & 6.00e-04 & 2B  \\
RWKV-7-488M            & 30         & 1024 & 4.00e-04 & 7B  \\
RWKV-7-1.5B            & 24         & 2048 & 4.00e-04 & 20B \\
\bottomrule
\end{tabularx}
\end{table*}

\begin{figure}[t]
  \centering
  \includegraphics[width=\linewidth, keepaspectratio=true]{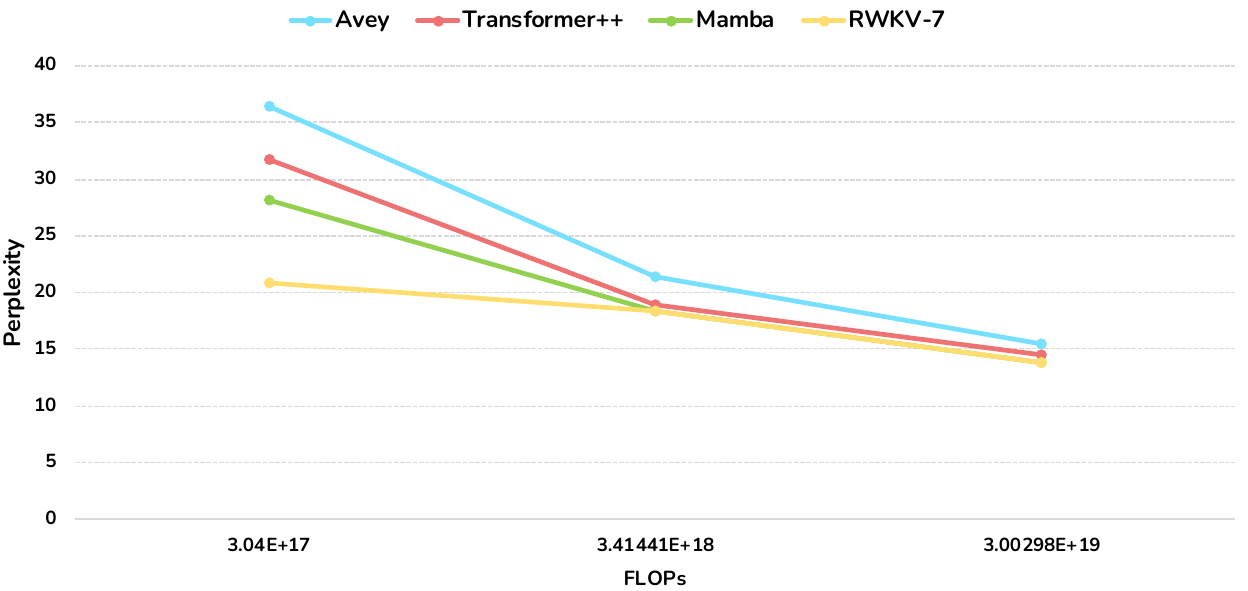}
  \caption{Scaling law results, comparing how perplexity decreases as compute increases, assuming three model sizes of 150M, 500M, and 1.5B parameters and a proportional increase in the number of training tokens with model size, following the Chinchilla scaling laws.}
  \label{fig:scaling_law_results}
\end{figure}

In this section, we present a scaling law study comparing how well Avey, Transformer++, Mamba, and RWKV-7 scale with increasing compute. For Avey, we use the full architecture, including both the ranker and neural processor. All models are trained at three different sizes, as defined in Section~\ref{sec:experimental_methodology} and summarized in Table~\ref{tab:scaling_config}. To ensure compute-optimal scaling, we proportionally increase the number of training tokens with model size, following the Chinchilla scaling laws~\citep{hoffmann2022training}. Specifically, and in line with the methodology from~\citep{gu2023mamba}, we use 2B, 7B, and 20B tokens to train models with approximately 150M, 500M, and 1.5B parameters, respectively. Lastly, we employ the same batch size across all the models to control for variability in the number of gradient update steps, especially because of training with a limited number of tokens.

Fig.~\ref{fig:scaling_law_results} presents the scaling results. The x-axis represents the total training compute budget, calculated as the product of the number of training tokens and model parameters, which serves as a proxy for the total FLOPs required to train each model. As shown, Avey exhibits the steepest decline in perplexity as compute increases. Although it begins with a relatively high perplexity\footnote{Avey can achieve substantially lower perplexity under alternative configurations. For example, the small model (150M parameters) with sequence length $N=1024$, split size $S=16$, and top-$k=15$ achieves much lower perplexity as shown in Table~\ref{tab:sensitivity_ranker}. The configuration used in this experiment-- and in Sections~\ref{sec:short_range} and~\ref{sec:long_range}, was selected based on its strong downstream benchmark performance, rather than optimal perplexity. As discussed in Appendix~\ref{sec:activation_enricher}, perplexity serves as a useful proxy for modeling capability, but does not always align perfectly with downstream task accuracy.}, it improves more rapidly than the other models, indicating strong scaling behavior and greater benefit from additional compute. Following Avey, Transformer++ demonstrates the next-best scaling trend, outpacing Mamba and RWKV-7. While Mamba achieves relatively low perplexity at smaller compute budgets, it does not scale as effectively as Avey or Transformer++. Finally, RWKV-7 performs well at low compute but shows the flattest scaling curve, suggesting it gains the least from additional training compute.


\section{Additional Short-Range Benchmark Results}
\label{sec:extra_short_range_results}

\begin{table*}[ht]
\centering
\caption{Performance of all models across short-range benchmarks at 90B, 95B, and 100B training tokens.}
\resizebox{\textwidth}{!}{%
\begin{tabular}{lcccccccc}
\toprule
\textbf{Model (\# of Tokens)} & \textbf{ARC-C} & \textbf{ARC-E} & \textbf{HellaSwag} & \textbf{PIQA} & \textbf{OBQA} & \textbf{SIQA} & \textbf{Winogrande} & \textbf{Avg.} \\
\midrule
Avey-153M (100BT) & 23.98 & 42.30 & 39.57 & 29.8 & 68.61 & 39.05 & 51.85 & 42.02 \\
Avey-153M (95BT)  & 24.23 & 42.09 & 39.21 & 31.2 & 68.23 & 39.15 & 50.28 & 42.06 \\
Avey-153M (90BT)  & 24.91 & 42.59 & 39.31 & 33.2 & 68.28 & 39.20 & 51.70 & 42.74 \\
\midrule
Transformer++-152M (100BT) & 23.29 & 43.43 & 39.47 & 29.4 & 67.03 & 39.10 & 50.51 & 41.90 \\
Transformer++-152M (95BT)  & 23.55 & 43.14 & 39.51 & 30.2 & 67.14 & 38.69 & 50.99 & 41.89 \\
Transformer++-152M (90BT)  & 24.06 & 42.93 & 38.97 & 29.8 & 66.87 & 38.89 & 49.17 & 41.24 \\
\midrule
Mamba-144M (100BT) & 24.32 & 43.73 & 40.82 & 29.8 & 68.28 & 39.00 & 52.41 & 42.62 \\
Mamba-144M (95BT)  & 23.63 & 43.69 & 40.51 & 32.2 & 68.06 & 39.82 & 53.35 & 43.61 \\
Mamba-144M (90BT)  & 24.57 & 43.18 & 40.33 & 29.2 & 68.61 & 39.41 & 52.41 & 42.53 \\
\midrule
RWKV-7-168M (100BT) & 23.89 & 43.14 & 41.50 & 29.8 & 68.72 & 39.41 & 50.99 & 42.35 \\
RWKV-7-168M (95BT)  & 24.23 & 42.89 & 41.77 & 29.2 & 68.99 & 39.10 & 51.14 & 42.48 \\
RWKV-7-168M (90BT)  & 24.40 & 43.01 & 41.38 & 30.0 & 68.44 & 39.00 & 51.14 & 42.48 \\
\midrule
\midrule
Avey-496M (100BT) & 27.13 & 48.99 & 52.17 & 32.0 & 72.47 & 40.53 & 54.54 & 46.55 \\
Avey-496M (95BT)  & 27.90 & 49.20 & 51.74 & 33.0 & 73.07 & 40.63 & 53.51 & 46.72 \\
Avey-496M (90BT)  & 27.47 & 48.65 & 51.56 & 32.4 & 71.93 & 39.30 & 55.09 & 46.63 \\
\midrule
Transformer++-488M (100BT) & 25.68 & 48.02 & 52.92 & 31.6 & 72.69 & 39.56 & 55.96 & 46.06 \\
Transformer++-488M (95BT)  & 27.39 & 47.90 & 52.69 & 31.6 & 72.36 & 40.07 & 54.22 & 46.12 \\
Transformer++-488M (90BT)  & 27.13 & 48.36 & 52.37 & 32.0 & 71.33 & 40.17 & 55.56 & 46.16 \\
\midrule
Mamba-500M (100BT) & 29.27 & 51.26 & 54.45 & 34.0 & 73.88 & 40.38 & 54.70 & 48.28 \\
Mamba-500M (95BT)  & 28.67 & 51.39 & 54.25 & 34.8 & 72.69 & 40.89 & 55.33 & 48.29 \\
Mamba-500M (90BT)  & 27.99 & 50.42 & 53.76 & 34.6 & 72.52 & 41.25 & 56.43 & 48.14 \\
\midrule
RWKV-7-501M (100BT) & 26.96 & 49.83 & 54.49 & 36.0 & 73.23 & 39.30 & 55.17 & 47.71 \\
RWKV-7-501M (95BT)  & 27.39 & 49.24 & 54.66 & 35.6 & 73.78 & 39.15 & 55.80 & 47.95 \\
RWKV-7-501M (90BT)  & 27.05 & 49.03 & 54.46 & 37.2 & 73.72 & 39.76 & 56.20 & 48.20 \\
\midrule
\midrule
Avey-1.52B (100BT) & 30.89 & 56.36 & 61.49 & 34.4 & 75.84 & 42.07 & 56.59 & 51.09 \\
Avey-1.52B (95BT)  & 32.34 & 56.94 & 61.63 & 37.6 & 75.57 & 41.76 & 58.09 & 52.42 \\
Avey-1.52B (90BT)  & 30.55 & 56.36 & 61.15 & 38.4 & 75.41 & 42.17 & 56.51 & 51.51 \\
\midrule
Transformer++-1.5B (100BT) & 30.29 & 56.19 & 64.28 & 38.8 & 76.12 & 42.27 & 61.33 & 52.75 \\
Transformer++-1.5B (95BT)  & 30.97 & 57.07 & 63.87 & 37.0 & 76.17 & 42.07 & 61.72 & 52.70 \\
Transformer++-1.5B (90BT)  & 28.75 & 55.60 & 63.45 & 38.2 & 75.73 & 42.37 & 61.09 & 52.19 \\
\midrule
Mamba-1.4B (100BT) & 32.42 & 57.87 & 64.78 & 38.4 & 76.61 & 42.48 & 62.27 & 53.55 \\
Mamba-1.4B (95BT)  & 32.85 & 57.91 & 64.37 & 35.4 & 76.33 & 42.02 & 60.93 & 52.69 \\
Mamba-1.4B (90BT)  & 32.00 & 58.63 & 64.38 & 36.8 & 76.22 & 41.50 & 61.33 & 52.69 \\
\midrule
RWKV-7-1.5B (100BT) & 32.42 & 59.55 & 64.59 & 37.4 & 76.82 & 41.86 & 59.67 & 53.19 \\
RWKV-7-1.5B (95BT)  & 33.11 & 58.88 & 64.49 & 37.0 & 76.88 & 41.71 & 60.38 & 53.21 \\
RWKV-7-1.5B (90BT)  & 33.28 & 58.71 & 64.21 & 37.0 & 76.82 & 41.56 & 60.14 & 53.10 \\
\bottomrule
\end{tabular}%
}
\label{tab:benchmark_results}
\end{table*}

\begin{table*}[ht]
\centering
\caption{Summary statistics for each model with different sizes computed over the last three checkpoints (i.e., at 90B, 95B, and 100B training tokens).}
\resizebox{\textwidth}{!}{%
\begin{tabular}{lcccc}
\toprule
\textbf{Model} & \textbf{Mean} & \textbf{Standard Deviation} & \textbf{Standard Error} & \textbf{95\% Confidence Interval} \\
\midrule
Avey-153M & 42.32 & 0.3683 & 0.2126 & (41.41, 43.24) \\
Transformer++-152M & 41.72 & 0.1821 & 0.1052 & (41.27, 42.17) \\
Mamba-144M & 42.73 & 0.2700 & 0.1559 & (42.06, 43.40) \\
RWKV-7-168M & 42.48 & 0.0094 & 0.0054 & (42.46, 42.51) \\
\midrule
Avey-496M & 46.82 & 0.1895 & 0.1094 & (46.35, 47.29) \\
Transformer++-488M & 46.65 & 0.0507 & 0.0293 & (46.52, 46.77) \\
Mamba-500M & 48.23 & 0.0835 & 0.0482 & (48.03, 48.44) \\
RWKV-7-501M & 48.00 & 0.1807 & 0.1043 & (47.55, 48.45) \\
\midrule
Avey-1.52B & 51.53 & 0.4497 & 0.2596 & (50.41, 52.65) \\
Transformer++-1.5B & 52.54 & 0.3218 & 0.1858 & (51.74, 53.34) \\
Mamba-1.4B & 53.12 & 0.3783 & 0.2184 & (52.18, 54.06) \\
RWKV-7-1.5B & 53.17 & 0.0553 & 0.0320 & (53.03, 53.30) \\
\bottomrule
\end{tabular}%
}
\label{tab:summary_statistics}
\end{table*}

To mitigate the effects of fluctuations in pre-training loss and downstream benchmark scores, we reported in Section~\ref{sec:short_range} average results across the final three checkpoints—taken at 5 billion token intervals (i.e., at 90B, 95B, and 100B tokens)—for all models evaluated, namely, Avey, Transformer++, Mamba, and RWKV-7. In this section, we provide the detailed performance scores for each model at each checkpoint in Table~\ref{tab:benchmark_results}. In addition, we summarize the mean, standard deviation, standard error, and 95\% confidence interval for each model, computed across the three checkpoints, in Table~\ref{tab:summary_statistics}. The illustrated statistical results reveal meaningful variance between models and across runs of the same model. For instance, while Mamba achieves the highest mean score of 42.73 among all the models in the small parameter regime ($\sim$150M parameters), it also exhibits a relatively wide confidence interval (42.06, 43.40) and a moderate standard deviation, highlighting nontrivial variability in performance across checkpoints.

Regarding variability across models, Table~\ref{tab:summary_statistics} shows overlapping confidence intervals, indicating that model rankings—particularly which model achieves the highest mean performance—could shift under minor experimental changes (e.g., random initialization, stochastic optimization, etc.). For example, in the small model regime, while Avey does not surpass Mamba in mean performance, their confidence intervals substantially overlap in the range (42.06, 43.24), suggesting that the two models are statistically comparable and that Avey could outperform Mamba in some runs. Similarly, a narrow but meaningful overlap exists between Avey and RWKV-7 in the range (42.46, 42.51), implying that Avey may occasionally match or slightly exceed RWKV-7 in certain cases. Lastly, although Mamba has the highest mean in this setting, its confidence interval also overlaps with RWKV-7, indicating that the difference in performance between the two models is not statistically significant and that RWKV-7 could match or slightly outperform Mamba in some runs.

In the medium model regime ($\sim$500M parameters), Avey outperforms Transformer++ in mean performance, but they are statistically comparable. In contrast, the performance gap between Avey and both Mamba and RWKV-7 is statistically significant at the 95\% confidence level, indicating that both models clearly outperform Avey in this setting. In the large model regime ($\sim$1.5B parameters), while Avey does not outpace Transformer++ in average performance, their confidence intervals overlap substantially, suggesting that Avey could potentially surpass Transformer++ in some runs. There is also a limited overlap between Avey and Mamba, indicating that while Mamba generally performs better, Avey might outperform it in certai n cases. In contrast, the difference between Avey and RWKV-7 is statistically significant at the 95\% level, confirming that RWKV-7 consistently outperforms Avey in this setting. Finally, although RWKV-7 has a slightly higher mean than Mamba (53.17 vs. 53.12), the meaningful overlap in their confidence intervals implies that the difference between them is not statistically significant, and either model could outperform the other depending on minor experimental factors.


\section{Additional Long-Range Benchmark Results}
\label{sec:extra_long_range_results}

\begin{figure}[ht]
  \centering
  \includegraphics[width=\linewidth, keepaspectratio=true]{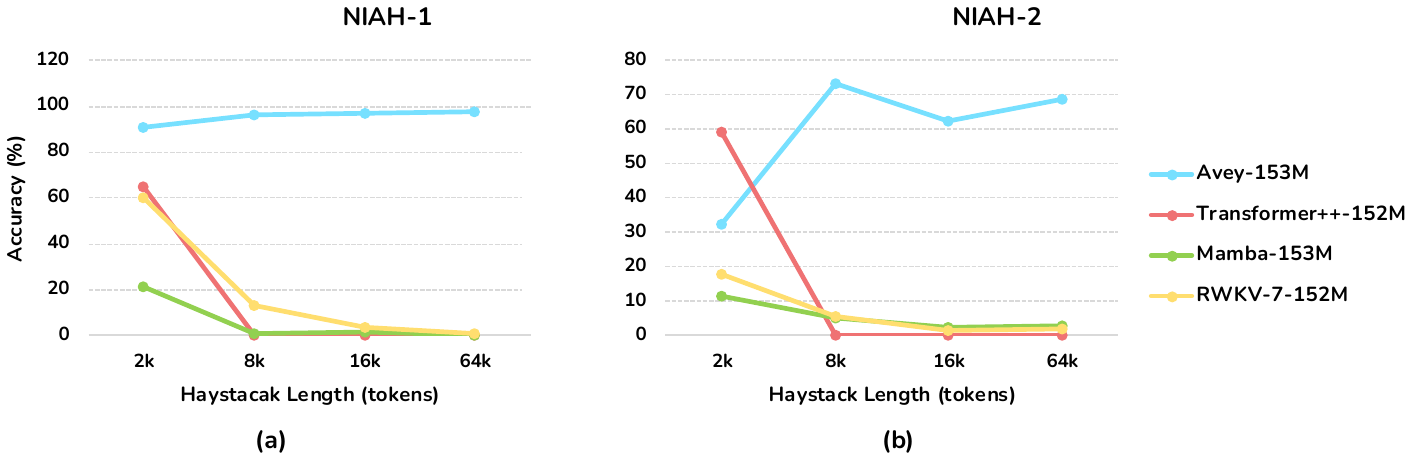}
  \caption{Performance comparison between Transformer++, Mamba, RWKV-7, and Avey on S-NIAH-1 and S-NIAH-2. The x-axis denotes the lengths of haystacks (i.e., documents with distractor texts, varying from 2k to 64k tokens). All models use  \textbf{$\sim$150M parameters}.}
  \label{fig:NiaH_1_and_2_small_models}
\end{figure}

\begin{figure}[t]
  \centering
  \includegraphics[width=\linewidth, keepaspectratio=true]{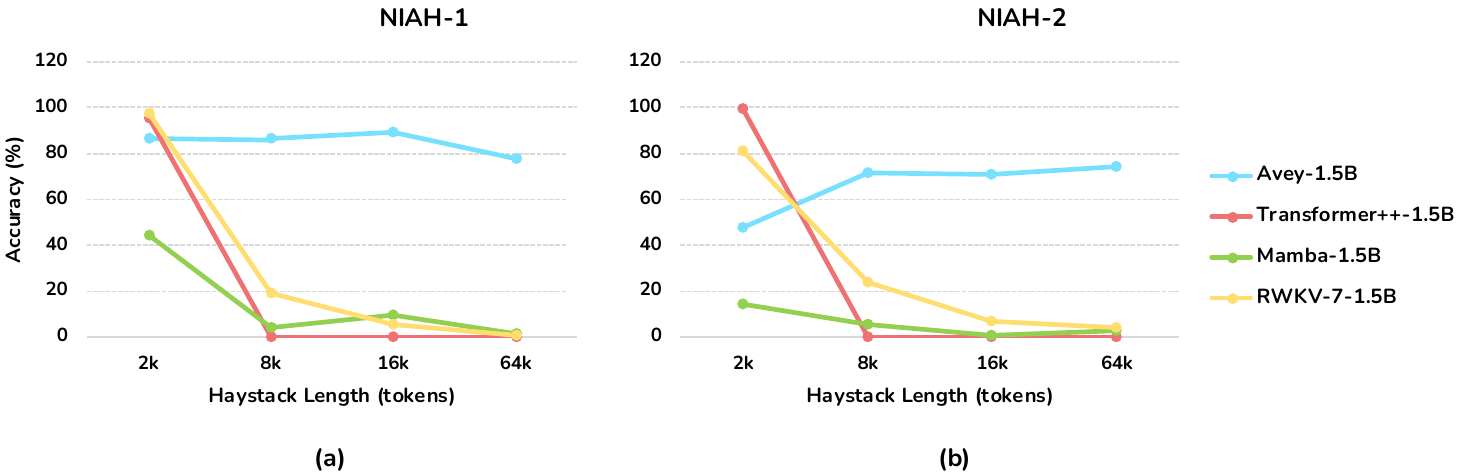}
  \caption{Performance comparison between Transformer++, Mamba, RWKV-7, and Avey on S-NIAH-1 and S-NIAH-2. The x-axis denotes the lengths of haystacks (i.e., documents with distractor texts, varying from 2k to 64k tokens). All models use  \textbf{$\sim$1.5B parameters}.}
  \label{fig:NiaH_1_and_2_large_models}
\end{figure}

In Section~\ref{sec:long_range}, we presented results for Avey, Transformer++, Mamba, and RWKV-7 under the medium parameter regime ($\sim$500M parameters) on the standard Single Needle-In-A-Haystack (S-NIAH) benchmark suite from RULER~\citep{hsieh2024ruler}, which is designed to evaluate models' abilities to handle long-range dependencies. The S-NIAH benchmark, along with two of its common variants—S-NIAH-1 and S-NIAH-2—was described in detail in Section~\ref{sec:long_range}. In this section, we extend our analysis by reporting results under two additional model regimes, small ($\sim$150M parameters) in Fig.~\ref{fig:NiaH_1_and_2_small_models} and large ($\sim$1.5B parameters) in Fig.~\ref{fig:NiaH_1_and_2_large_models}. Akin to the experiment in Section~\ref{sec:long_range}, Transformer++, Mamba, and RWKV-7 were trained with a context window of 2048 tokens, while Avey was trained with a shorter window of only 512 tokens.

In both the small and large model regimes, under S-NIAH-1 and S-NIAH-2, Transformer++, Mamba, and RWKV-7 perform well when the haystack length is 2k, fitting within their trained context windows. Yet, Mamba consistently underperforms Transformer++ and RWKV-7, likely due to solely relying on recurrence, which somehow treats the entire input uniformly, making the model more susceptible to distractions from irrelevant tokens. In contrast, RWKV-7, which combines recurrence with attention, performs better than Mamba but remains inferior to Transformer++, potentially because the attention mechanism allows it to prioritize tokens relevant to the needle, while the recurrent component may still contribute to signal dilution. Transformer++, relying exclusively on full attention, achieves the best performance within the context window by effectively focusing on relevant tokens without interference from recurrence-based mechanisms. Nonetheless, once the haystack length exceeds the models’ context windows, all the three models exhibit a substantial drop in performance. Mamba and RWKV-7, however, show minimal generalization beyond their training limits compared to Transformer++, as previously discussed in Section~\ref{sec:long_range}.

Compared to Transformer++, Mamba, and RWKV-7, Avey generalizes far beyond its trained context window on both S-NIAH-1 and S-NIAH-2 across all parameter regimes, underscoring its strong extrapolative capabilities (as also shown in Section~\ref{sec:long_range}). Notably, this holds despite Avey being trained with a context window of only 512 tokens. For example, in the small parameter regime, Avey achieves an accuracy of 97.8\% on S-NIAH-1 with a 64k-token haystack, while Transformer++, Mamba, and RWKV-7 drop to 0\%, 0\%, and 0.6\%, respectively. Similarly, on S-NIAH-2 at the same haystack length, Avey attains 68.8\% accuracy, whereas Transformer++, Mamba, and RWKV-7 fall to 0\%, 2.8\%, and 2\%, respectively. Comparable trends are observed in the large parameter regime as well, as illustrated in Fig.~\ref{fig:NiaH_1_and_2_large_models}.

An interesting observation arises in the small parameter regime, where Avey outperforms all other models on S-NIAH-1 with a haystack length of 2k, knowing that this length exceeds its trained context window width and enables it to demonstrate its strong extrapolative capability. However, this pattern does not persist in the medium (see Fig.~\ref{fig:NiaH_1_and_2} in Section~\ref{sec:long_range}) and large (see Fig.~\ref{fig:NiaH_1_and_2_large_models}) parameter regimes, where Transformer++ and RWKV-7 outperform Avey on the same benchmark at 2k length, despite this length still surpassing Avey’s trained context window. This suggests that these models, with their increased capacity, are able to compensate for the challenge posed by S-NIAH-1, and entails that Avey might benefit from a longer training context window. 

In this paper, we kept Avey’s context window fixed at 512 tokens across all parameter regimes. All tuning experiments related to sequence length, split size, and top $k$ splits (see Section~\ref{sec:sensitivity_ranker}) were conducted exclusively using the small model size. It is plausible that with larger capacity, Avey could more effectively leverage longer sequences by retrieving and contextualizing a larger set of relevant tokens while filtering out less informative ones, thereby enhancing contextual representations and further boosting performance. Investigating the relationship between sequence length and model size in Avey is an interesting direction for future work.

\section{Complexity Analysis}
\label{sec:complexity_analysis}


As indicated in Sections~\ref{sec:ranker}, \ref{enricher}, \ref{contextualizer}, and \ref{fuser}, the training time complexities of the ranker, enricher, contextualizer, and fuser are \( \mathcal{O}(N^2 d) \), \( \mathcal{O}(N m d) \), \( \mathcal{O}(N k C m_t) \), and \( \mathcal{O}(N m d) \), respectively, where \( N \) is the sequence length, \( S \) is the split size, \( d \) is the original embedding dimension, \( m \) is the projected embedding dimension (with \( m > d \)), \( m_t \) is the tail part of \( m \) forwarded to the contextualizer, \( C \) is the context width (with \( C \leq N \)), and \( k \) is the number of splits contextualized with each current split. This yielded an overall training time complexity of \( \mathcal{O}(N^2 d) \), assuming that scalar multiply-add operations (e.g., those used in computing cosine similarity for MaxSim) and comparisons (e.g., those used to determine maximum scores) are constant-time.

During inference, the time complexities of the enricher, contextualizer, and fuser remain unchanged. However, the ranker's analysis slightly changes, as at {\em each} time step \( t \) (i.e., upon predicting a new token), the current split is compared against all previous splits. More precisely, at each time step \( t \), the current split—denoted as split \( i \) and incrementally filled as tokens are generated—is compared against all \( i - 1 \) preceding splits, each consisting of \( S \) tokens.  Consequently, the cost of comparing \( t \) tokens in split \( i \) (with \( t \leq S \)) against \( S \) tokens in a previous split is \( \mathcal{O}(t \cdot S \cdot d) \).

Now, if we let \( M = \lceil N / S \rceil \approx N / S \) for large \( N \) be the number of splits, the total inference cost can be defined as:

\[
\sum_{i=1}^{M} \sum_{t=1}^{S} (i - 1) \cdot \mathcal{O}(t \cdot S \cdot d)
\]

Simplifying the inner summation yields:

\[
\sum_{t=1}^{S} (i - 1) \cdot \mathcal{O}(t \cdot S \cdot d)
= (i - 1) \cdot \mathcal{O}(S \cdot d) \cdot \sum_{t=1}^{S} t
= (i - 1) \cdot \mathcal{O}(S \cdot d) \cdot \frac{S(S + 1)}{2}
= (i - 1) \cdot \mathcal{O}(S^2 \cdot d)
\]

Substituting this back into the outer summation gives:

\[
\sum_{i=1}^{M} (i - 1) \cdot \mathcal{O}(S^2 \cdot d)
= \mathcal{O}(S^2 \cdot d) \cdot \sum_{i=1}^{M} (i - 1)
= \mathcal{O}(S^2 \cdot d) \cdot \frac{M(M - 1)}{2}
\]

Substituting \( M \) with \( N / S \) for large \( N \) results in:

\[
\mathcal{O}(S^2 \cdot d) \cdot \frac{(N/S)^2}{2}
= \mathcal{O}(S^2 \cdot d) \cdot \mathcal{O}(N^2 / S^2)
= \mathcal{O}(N^2 \cdot d)
\]

Therefore, the cost per token becomes:
\[
\frac{\mathcal{O}(N^2 d)}{N} = \mathcal{O}(N d)
\]

\section{Is the Ranker a RAG Component?}
\label{sec:ranker_and_rag}

The ranker is an {\em internal} component of Avey that operates \emph{within} the input sequence, selecting among its splits for more effective contextualization. It does not query external corpora or indexes, introduces no retrieval I/O or freshness dependencies, and adds no retrieval latency. Its role is architectural, that is, to allocate Avey’s internal contextual budget and decouple context width from sequence length so that Avey can fully contextualize sequences far beyond its training window.

By contrast, Retrieval\mbox{-}Augmented Generation (RAG)~\citep{lewis2020rag} augments a model with \emph{external} (non\mbox{-}parametric) knowledge via a retriever, ranging from classic BM25~\citep{robertson2009bm25} and dense passage retrieval (DPR)~\citep{karpukhin2020dpr} to system\mbox{-}level designs such as REALM~\citep{guu2020realm}, RETRO~\citep{borgeaud2022retro}, and MacRAG~\citep{lim2025macrag}, among others. RAG aims to: (1) improve factuality by grounding outputs in retrieved documents, (2) make models updatable by reflecting new information without retraining, and (3) reduce compute for long\mbox{-}context tasks by moving knowledge out of weights.

As such, the two mechanisms are \emph{orthogonal}. The ranker allocates the model’s internal contextual budget over the given sequence, whereas RAG changes the evidence set by importing out\mbox{-}of\mbox{-}sequence content. They can be composed (RAG can be layered atop Avey, as it is with the Transformer) but one does not subsume the other.

\section{Neural Contextualization vs. Attention}
\label{sec:contextualization_attention}

Avey’s contextualizer is an embedding-wise neural network that dispenses with attention. In Appendix~\ref{sec:ablation}, we replaced it with standard self-attention and observed a \(4.6\%\) increase in perplexity alongside a \(2.1\%\) decline in average task performance, underscoring its central role in Avey’s architecture.

Formally, the contextualizer is defined in Equation~\ref{eq:contextualizer}, repeated below for convenience:
\[
\mathbf{c}(\mathbf{Z}_t)
= \mathbf{Z}_{tl}\ \odot\
\sigma\!\Big(\big(\mathbf{V}\ \odot\ \mathcal{N}(\mathbf{Z}_{tr})\,\mathcal{N}(\mathbf{Z}_{tr})^\top\big)\,\mathbf{Z}_{tr}\ +\ \mathbf{b}'\Big).
\]

Let \(\mathbf{S} := \mathcal{N}(\mathbf{Z}_{tr})\,\mathcal{N}(\mathbf{Z}_{tr})^\top\). The product \((\mathbf{V}\!~\odot\!~\mathbf{S})\mathbf{Z}_{tr}\) yields a content-dependent signal that is passed through a pointwise nonlinearity and used to gate \(\mathbf{Z}_{tl}\) elementwise, producing a bounded, feature-wise modulation rather than a mixture over values. By contrast, self-attention computes a row-stochastic convex combination of value vectors after a Q/K split and softmax normalization. Evidently, our formulation departs away from both the softmax and the Q/K/V decomposition, whereby weights are neither constrained to be nonnegative nor to sum to one, and the output acts as a gate on carrier features (i.e., \(\mathbf{Z}_{tl}\)) rather than a convex average of value vectors.

The contextualizer also differs fundamentally from linear attention~\citep{choromanski2021performer, katharopoulos2020transformers, wang2020linformer, beltagy2020longformer, sun2023retentive}. Linear-attention variants obtain near-linear complexity by exploiting an associative kernel factorization that permits reordering and prefix accumulation, typically of the form \(\phi(\mathbf{Q})\big(\phi(\mathbf{K})^\top \mathbf{V}_{\!\text{val}}\big)\). Equation~\ref{eq:contextualizer} does not admit such reordering. In fact, the Hadamard coupling (\(\mathbf{V}\!~\odot\!~\mathbf{S}\)) breaks the algebraic associativity required to push multiplications across terms, and the normalization \(\mathcal{N}(\cdot)\) is neither linear nor guaranteed nonnegative, precluding the kernel tricks used to approximate softmax attention with associative feature map functions (e.g., ReLU and Exp). Lastly, we note that the contextualizer remains quadratic (not linear) in sequence length.

For similar reasons, Equation~\ref{eq:contextualizer} cannot be reformulated as a finite-state RNN under an autoregressive mask. Let \(\mathbf{S}_t=\mathcal{N}(\mathbf{Z}_{tr}^{\le t})\,\mathcal{N}(\mathbf{Z}_{tr}^{\le t})^\top\). The update at step \(t\!+\!1\) depends on the full pairwise matrix \((\mathbf{V}\!~\odot\!~\mathbf{S}_t)\), that is, on all position-specific interactions among the past tokens after data-dependent normalization. Because the learned weight matrix \(\mathbf{V}\) introduces position-dependent multiplicative couplings, there is no time-invariant transition \(h_{t+1}=f(h_t,x_{t+1})\) with a fixed-dimensional sufficient statistic \(h_t\) that exactly summarizes \((\mathbf{V}\!~\odot\!~\mathbf{S}_t)\). In particular, the required weights vary across positions and must be recomputed, so any streaming recurrence would either approximate by tying/averaging \(\mathbf{V}\) or maintain \(\mathcal{O}(t)\) state. Therefore, an exact finite-state RNN equivalence is unavailable.


Empirically, \(\mathbf{V}\) performs most of the heavy-lifting in Avey, while \(\mathbf{S}\) primarily induces {\em selectivity}, dynamically emphasizing or suppressing interactions conditioned on the input, echoing the selectivity principle advocated in recent sequence models~\citep{gu2023mamba}. An ablation in Appendix~\ref{sec:ablation} shows that including \(\mathbf{S}\) delivers a consistent, albeit modest, gain by making the neural processor’s parametrization input-adaptive. 

Putting everything together, these distinctions (i.e., gating rather than mixing, non-associative pairwise modulation rather than kernel-factorizable operations, and explicit quadratic interactions), explain both the theoretical departure from self-attention and linear attention and the observed empirical contribution of the contextualizer within Avey.

\section{Design Rationale}
\label{sec:design_rationale}

We designed Avey around clear functional roles for its core modules. Below, we outline some of the guiding intuitions and how they inform its architecture.

\paragraph{Enricher:}
A substantial body of evidence indicates that much of a language model’s knowledge is stored in feed\mbox{-}forward sublayers and accessed through non\mbox{-}linear feature interactions~(e.g., \citep{geva2021transformer_ffn_memory}). The \emph{enricher} is designed accordingly. It serves both as the primary repository of parametric knowledge and as a mechanism for intra\mbox{-}embedding interactions, enabling higher\mbox{-}order, non\mbox{-}linear composition of features within each embedding. This improves expressivity by allowing features to modulate and refine one another in a context\mbox{-}aware manner.

\paragraph{Contextualizer:}
The \emph{contextualizer} operates as an embedding\mbox{-}wise neural network such that each neuron forms a weighted sum over input embeddings with learned coefficients (see Equation~\ref{eq:contextualizer}). To introduce input\mbox{-}dependent {\em selectivity} (as in~\citep{gu2023mamba}), we augment these static weights with a cosine\mbox{-}similarity term that produces a second, data\mbox{-}driven set of weights (the two are combined multiplicatively via a Hadamard product). This dynamic modulation improves behaviors such as copying and induction by strengthening interactions that are semantically relevant to the current input. The split\mbox{-}and\mbox{-}gate structure follows established gated designs in gMLP~\citep{liu2021pay} and GLU variants~\citep{dauphin2017language,shazeer2020glu,wu2019pay}.

\paragraph{Partial Embedding Bypassing:}
The enricher’s output is partitioned into two streams, one is passed to the contextualizer and the other is bypassed and fed directly to the {\em fuser}. The bypassed part plays two complementary roles. First, it provides a strong residual path that preserves signal and stabilizes optimization by improving gradient flow within each Avey layer. Second, it supplies additional non\mbox{-}linear capacity in the downstream feed\mbox{-}forward fuser, complementing the contextualizer’s primarily linear mixing across embeddings. This balance between context\mbox{-}aware and context\mbox{-}invariant processing yields richer, more diverse representations.

\paragraph{Fuser:}
The \emph{fuser} (a position\mbox{-}wise feed\mbox{-}forward network) learns how to combine the contextualized and bypassed streams and then projects the result back to the model’s embedding dimension, ensuring compatibility with residual pathways across layers. As a feed\mbox{-}forward network (FFN), it also contributes to storing and accessing parametric knowledge learned during training, analogous to FFN roles in Transformers.

\section{Limitations}
\label{sec:scope_and_limitations}

The scope of this work is limited to textual data and does not extend to other modalities such as images or audio, nor to specialized domains like genomics. Our evaluation of Avey focuses on standard autoregressive language modeling, benchmarking against popular open-source architectures using both pretraining metrics (perplexity) and zero-shot performance on established NLP benchmarks. Consequently, we do not investigate Avey’s ability to produce bidirectional contextualized representations as in BERT~\citep{devlin2019bert}. We leave this direction to future work. Finally, the paper emphasizes effectiveness rather than efficiency. While our complexity analysis indicates quadratic training time similar to Transformers, our current implementation is slower in practice, and additional engineering is needed to optimize it.


\end{document}

%% file: math_commands.tex
\usepackage{amsmath,amsfonts,bm}

\def\eqref#1{equation~\ref{#1}}

\def\1{\bm{1}}

\DeclareMathAlphabet{\mathsfit}{\encodingdefault}{\sfdefault}{m}{sl}
\SetMathAlphabet{\mathsfit}{bold}{\encodingdefault}{\sfdefault}{bx}{n}

